%% file: main.tex
\documentclass{article}
\usepackage{kr}

\usepackage{times}
\usepackage{soul}
\usepackage{url}
\usepackage[hidelinks]{hyperref}
\usepackage[utf8]{inputenc}
\usepackage[small]{caption}
\usepackage{graphicx}
\usepackage{amsmath}
\usepackage{amsthm}
\usepackage{booktabs}
\usepackage{algorithm}
\usepackage{algorithmic}
\usepackage{natbib}

\input{preamble}
\usepackage[capitalize,noabbrev]{cleveref}

\AddToHook{env/plemma/begin}{\crefalias{ptheorem}{plemma}}

\crefname{ptheorem}{THEOREM}{THEOREMS}
\crefname{plemma}{lemma}{lemmas}
\Crefname{ALC@unique}{Line}{Lines}
\newcounter{myalg}
\AtBeginEnvironment{algorithmic}{\refstepcounter{myalg}}
\makeatletter
\@addtoreset{ALC@unique}{myalg}
\makeatother

\newcommand{\resnet}{ResNet50\xspace}
\newcommand{\swint}{Swin\_t\xspace}
\newcommand{\mob}{MobileNet\xspace}
\newcommand{\voc}{PascalVOC\xspace}
\newcommand{\ecssd}{ECSSD\xspace}

\newcommand{\onecomplete}{$1$-complete\xspace}
\newcommand{\deltacomplete}{$\delta$-complete\xspace}

\usepackage{latexsym}

\title{Sufficient, Necessary and Complete Causal Explanations\\in Image Classification}

\author{%
David A. Kelly$^1$,
Hana Chockler$^1$,
\affiliations
$^1$ King's College London, UK\\
\emails
\{david.a.kelly, hana.chockler\}@kcl.ac.uk,
}

\begin{document}

\maketitle

\begin{abstract}
Existing algorithms for explaining the outputs of image classifiers are based on a variety of approaches and produce
explanations that frequently lack formal rigour. On the other hand, logic-based explanations are formally and rigorously defined but 
their computability relies on strict assumptions about the model that do not hold on image classifiers.

In this paper, we show that causal explanations, in addition to being formally and rigorously defined, 
enjoy the same formal properties as logic-based ones, while still lending themselves
to black-box algorithms and being a natural fit for image classifiers. We prove formal properties of causal
explanations and their equivalence to logic-based explanations. We demonstrate how to subdivide an image into its sufficient and necessary components. We introduce \deltacomplete explanations, which have a minimum confidence threshold and \onecomplete causal explanations, explanations that are classified with the same confidence as the original image.

We implement our definitions, and our experimental results demonstrate that different models have different patterns of sufficiency, 
necessity, and completeness. Our algorithms are efficiently computable, 
taking on average $6$s per image on a \resnet model to compute all types of explanations, and are totally black-box, needing no knowledge of the model, no access to model internals, no access to gradient, nor requiring any properties, such as monotonicity, of the model.
\end{abstract}

\section{Introduction}\label{sec:intro}

\begin{figure*}[t]
    \centering
    \begin{subfigure}[t]{0.18\textwidth}
        \centering
        \includegraphics[scale=0.3]{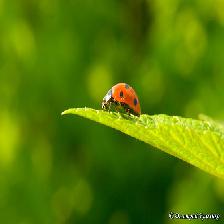}
        \caption{Ladybug}%
        \label{subfig:lady}
    \end{subfigure}
    \hfill
    \begin{subfigure}[t]{0.18\textwidth}
        \centering
        \includegraphics[viewport=90 85 160 150,clip]{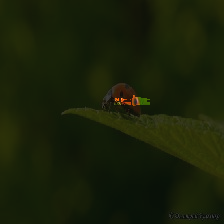}
        \caption{Sufficient explanation}%
        \label{subfig:lady_suff}
    \end{subfigure}
    \hfill
    \begin{subfigure}[t]{0.18\textwidth}
        \centering
        \includegraphics[viewport=90 85 160 150,clip]{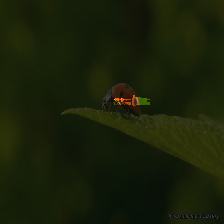}
        \caption{adolescent newt (eft)}%
        \label{subfig:lady_21}
    \end{subfigure}
    \hfill
    \begin{subfigure}[t]{0.18\textwidth}
        \centering
        \includegraphics[scale=0.3]{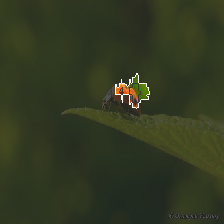}
        \caption{\deltacomplete explanation}%
        \label{subfig:lady_con}
    \end{subfigure}
    \hfill
    \begin{subfigure}[t]{0.18\textwidth}
        \centering
        \includegraphics[scale=0.3]{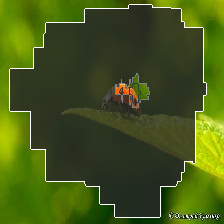}
        \caption{\onecomplete explanation}%
        \label{subfig:complete}
    \end{subfigure}
    \caption{$4$ types of explanation for ``ladybug'' with $0.46$ confidence on a \resnet. ~\Cref{subfig:lady_suff} shows a subset of pixels sufficient to obtain class ``ladybug''.\Cref{subfig:lady_21} shows that adding just $10$ more pixels to~\Cref{subfig:lady_suff} changes the classification. In this paper we introduce `complete' explanations, which are subsets of pixels that are sufficient and necessary for ``ladybug'', and removing these pixels results in ``leaf beetle'' (\Cref{subfig:lady_con}), and \onecomplete explanations (\Cref{subfig:complete}), which are subsets of pixels which are complete and have the original confidence of $0.46$.}%
    \label{fig:ladybird}
\end{figure*}

Recent progress in artificial intelligence and the ever increasing deployment of AI systems has highlighted the need to understand better \emph{why} some decisions are made by such systems and \emph{what} information they are using. For example, one may need to know why a classifier decides that an MRI scan showed evidence of a tumor~\citep{blake2024explainable}. Answering such questions is the province of causality. A causal explanation for an image classification is a special case of explanations in actual causality~\citep{Hal19} and 
identifies a minimal set of pixels which, by themselves, are sufficient to re-create the original top-$1$~\citep{CH24}. 

Logic-based explanation approaches provide formal guarantees, but their framework assumes that the model is given explicitly as a function. Their \emph{formal abductive explanations}, or \emph{prime implicants} (PI), are defined as
sets of features such that, if they take the given values, always lead to the same decision~\citep{shih2018}. 
Logic-based methods can also compute \emph{contrastive} explanations, that is, those features which, if altered, change the original decision. These abductive and contrastive explanations require a model to be monotonic or linear to be effectively computable~\citep{marques21a},
and therefore are not suitable for image classifiers.

In this paper, we show that causal explanations enjoy all the formal properties of logic-based explanations, while not putting any
restrictions on the model and being efficiently computable for black-box image classifiers.
We prove that a causal explanation in our setting is equivalent to an abductive explanation. Furthermore, \textbf{we augment the actual causality framework with the model's confidence in the classification, introducing $\delta$-confident explanations, and use these to produce more fine-grained and robust results. We show how to calculate a complete (sufficient and necessary) subset of an image.} By masking the complete pixels, we can calculate the \textbf{inverse classification}, which characterises the remaining data in the image. We also introduce a causal version of the `completeness`
property for explanations, following~\cite{srinivas19}, which we call \textbf{$1$-completeness; a complete explanation with the same score as the original image.}
In addition to the inverse classification, we explore the pixel differences between a
$\delta$-complete explanation and a \onecomplete, which we call the \textbf{adjustment pixels}. 

We examine the relationship between \deltacomplete and adjustment pixel sets by examining the semantic distance between the original classification and the classifications of these pixel sets, as illustrated in \Cref{fig:ladybird}. \textbf{Our approach allows us to formally subdivide an image into its sufficient pixels, 
complete pixels and adjustment pixels. This can be done for any confidence threshold. We know of no other method which achieves this for image classifiers.}
We prove complexity results for our definitions, giving a justification to efficient approximation algorithms.

Our algorithms are based on \rex~\citep{CKKS24}. In~\Cref{sec:algo}, we introduce black-box approximation algorithms to compute
\deltacomplete and \onecomplete causal explanations for image classifiers.
Our algorithms do not require any knowledge of model architecture, no access to the model internals, nor do they require any specific properties of the model. We implemented our algorithms and present experimental results on three state-of-the-art models and three standard benchmark datasets. \textbf{We apply our definitions and algorithms to two other \xai tools to demonstrate the utility of the method while highlighting the benefits of applied causal reasoning.}

\paragraph{Explainability vs. Interpretability} We do not intend to imply that causal explanations are more interpretable than other forms of explanation. As \cite{bhusalface} point out: `models are not constrained to use human-understandable cues; they only use features that minimize loss'. Indeed, it often the case that causal explanations are very small~\citep{Kelly25}, but they are sufficient to elicit the desired model behavior.
We are interested in the formal partitioning of pixels in an image into different functional sets. These functional sets reveal important information about the inner workings of the model. Due to the lack of space, proofs and some evaluation results are deferred to the supplementary material.

\section{Background}\label{sec:suff}
There are many different definitions of explanation in the \xai literature; some are saliency-based~\citep{CAM}, some are gradient-based~\citep{srinivas19}, Shapley-based~\citep{gradientshap} or train locally interpretable models~\citep{lime}. 
Logic-based explanations are different in having a mathematically precise definition. Causal explanations, as defined in~\cite{CH24} and \cite{CKKS24} have much more in common with the rigorous definitions of logic-based explanations. 

\begin{figure*}[t]
    \centering
    \begin{subfigure}[t]{0.18\textwidth}
        \centering
        \includegraphics[scale=0.21]{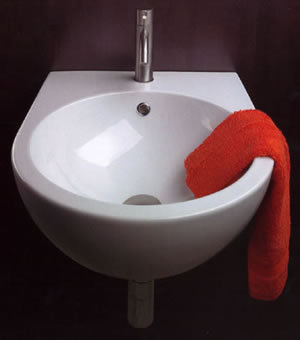}
        \caption{Wash basin/sink ($0.6$ confidence)}%
        \label{fig:sink}
    \end{subfigure}
    \hfill
    \begin{subfigure}[t]{0.18\textwidth}
        \centering
        \includegraphics[scale=0.3]{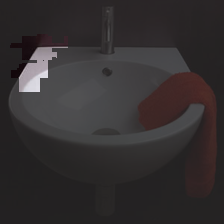}
        \caption{Sufficient explanation ($0.04$ confidence)}%
        \label{fig:sink_suff}
    \end{subfigure}
    \hfill
    \begin{subfigure}[t]{0.18\textwidth}
        \centering
        \includegraphics[scale=0.3]{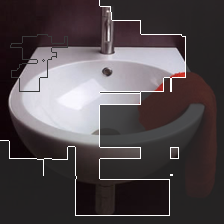}
        \caption{\deltacomplete explanation $(0.757$ confidence)}%
        \label{fig:sink_con}
    \end{subfigure}
    \hfill
    \begin{subfigure}[t]{0.18\textwidth}
        \centering
        \includegraphics[scale=0.3]{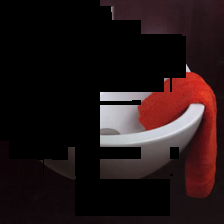}
        \caption{\deltacomplete explanation mask}%
        \label{fig:sink_inverse}
    \end{subfigure}
    \hfill
    \begin{subfigure}[t]{0.18\textwidth}
        \centering
        \includegraphics[scale=0.41]{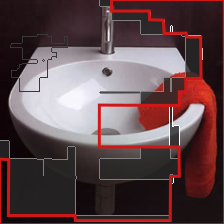}
        \caption{\onecomplete explanation (adjustment pixels in red)}%
        \label{fig:sink_com}
    \end{subfigure}
    \caption{A `washbasin' partitioned into sufficient, \deltacomplete, and adjustment pixel sets. $1$-completeness required $82\%$ of the image for a \resnet model. The sufficient set, \Cref{fig:sink_suff}, is very small, with low confidence. The \deltacomplete explanation (\Cref{fig:sink_con}) has higher confidence than the original image. Masking out~\Cref{fig:sink_con} to get~\Cref{fig:sink_inverse}, \resnet gives us a classification of `toilet seat'. Interestingly, the adjustment pixels (\Cref{fig:sink_com}) reduce model confidence from $0.75$ to $0.6$, even though the they are also classified as `wash basin'.}
    \label{fig:complete_example}
\end{figure*}

\subsection{Actual causality}\label{subsec:cause} 
In what follows, we briefly introduce the relevant definitions from the theory of actual causality. The reader is referred to~\cite{Hal19} for further reading. 
We assume that the world is described in terms of variables and their values.  
Some variables may have a causal influence on others. This influence is modeled by a set of {\em structural equations}.
It is conceptually useful to split the variables into two
sets: the {\em exogenous\/} variables $\U$, whose values are
determined by factors outside the model, and the {\em endogenous\/} variables $\V$, whose values are ultimately determined by the exogenous variables.  
The structural equations $\cF$ describe how these values are 
determined. A \emph{causal model}, $M$, is described by its variables and the structural
equations. We restrict the discussion to acyclic (recursive) causal models.
A \emph{context}, $\vec{u}$, is a setting for the exogenous variables $\U$, which then
determines the values of all other variables. 
We call a pair $(M,\vec{u})$, consisting of a causal model $M$ and a
context $\vec{u}$, a \emph{(causal) setting}.
An intervention is defined as setting the value of some
variable $X$ to $x$, and essentially amounts to replacing the equation for $X$
in $\cF$ by $X = x$. 
%A \emph{probabilistic causal model} is a pair $(M,\Pr)$, where $\Pr$ is a probability distribution on contexts.

A causal formula $\psi$ is true or false in a setting.
We write $(M,\vec{u}) \sat \psi$  if
the causal formula $\psi$ is true in
the setting $(M,\vec{u})$.
Finally, 
$(M,\vec{u}) \sat [\vec{Y} \gets \vec{y}]\varphi$ if 
$(M_{\vec{Y} = \vec{y}},\vec{u}) \sat \varphi$,
where $M_{\vec{Y}\gets \vec{y}}$ is the causal model that is identical
to $M$, except that the 
variables in $\vec{Y}$ are set to $Y = y$
for each $Y \in \vec{Y}$ and its corresponding 
value $y \in \vec{y}$.

A standard use of causal models is to define \emph{actual causation}: that is, 
what it means for some particular event that occurred to cause 
another particular event. 
There have been a number of definitions of actual causation given
for acyclic models
\citep{beckers21c,GW07,Hall07,HP01b,Hal19,hitchcock:99,Hitchcock07,Weslake11,Woodward03}.
In this paper, we focus on what has become known as the \emph{modified} 
Halpern--Pearl (HP) definition and some related definitions introduced
in~\cite{Hal19}. The events that can be causes are arbitrary conjunctions of primitive
events. %(formulae of the form $X=x$). 

\dfn[Actual cause\label{def:AC}]
$\vec{X} = \vec{x}$ is 
an \emph{actual cause} of $\varphi$ in $(M,\vec{u})$ if the
following three conditions hold: 
\begin{description}
\item[{\rm AC1.}]\label{ac1} $(M,\vec{u}) \models (\vec{X} = \vec{x})$ and $(M,\vec{u}) \models \varphi$. 
\item[{\rm AC2}]\label{ac2} There is a (possibly empty) set $\vec{W}$ of variables in $\vec{V}$ and a setting $\vec{x}'$
of the variables in $\vec{X}$ such that if $(M, \vec{u}) \models \vec{W} = \vec{w}^*$, then
$(M, \vec{u}) \models [\vec{X} \leftarrow \vec{x}', \vec{W} \leftarrow \vec{w}^*]\neg{\varphi}$
\item[{\rm AC3.}] \label{ac3}\index{AC3}  
  $\vec{X}$ is minimal; there is no strict subset $\vec{X}'$ of
  $\vec{X}$ such that $\vec{X}' = \vec{x}''$ can replace $\vec{X} =
  \vec{x}'$ in 
  AC2, where $\vec{x}''$ is the restriction of
$\vec{x}'$ to the variables in $\vec{X}'$.
\end{description}
\edfn
\noindent 
We can show the counterfactual dependence of $\varphi$ on $\vec{X}$ by holding the variables in $\vec{W}$ at their actual values. In the special case that $\vec{W} = \emptyset$, we get the 
but-for definition. A variable $x$ in an actual cause $\vec{X}$ is called a \emph{part of a cause}. In what follows, we adopt the convention of Halpern and state that \emph{part of a cause is a cause}.

The notion of explanation taken from~\citet{Hal19} is relative to a set of contexts.
\dfn[Explanation\label{def:EX}]
$\vec{X} = \vec{x}$ is 
an \emph{explanation} of $\varphi$ relative to a set $\K$ of contexts 
in a causal model $M$ if the following conditions hold:  
\begin{description}
\item[{\rm EX1a.}]  
If $\vec{u} \in \K$ and $(M,\vec{u}) \models (\vec{X} = \vec{x})
  \wedge \varphi$, then there exists a conjunct $X=x$ of $\vec{X} =
  \vec{x}$ and a (possibly empty) conjunction $\vec{Y} = \vec{y}$ such
  that $X=x \wedge \vec{Y} = \vec{y}$ is an actual cause of $\varphi$
  in $(M,\vec{u})$. 
\item[{\rm EX1b.}] $(M,\vec{u}') \models [\vec{X} \leftarrow \vec{x}]\varphi$  for all
  contexts $\vec{
    u}' \in \K$. 
\item[{\rm EX2.}] $\vec{X}$ is minimal; there is no
  strict subset $\vec{X}'$ of $\vec{X}$ such that $\vec{X}' =
  \vec{x}'$ satisfies EX1,  
where $\vec{x}'$ is the restriction of $\vec{x}$ to the variables in $\vec{X}'$.
\item[{\rm EX3.}] \label{ex3} $(M,u) \sat \vec{X} = \vec{x} \wedge
  \varphi$ for some $u \in \K$.
\end{description}
\edfn

\subsection{Actual causality in image classification}\label{subsec:cause-img}
The material here is taken from~\cite{CKKS24}, and the reader is referred to 
this paper for a complete overview of causal models for black-box image classifiers.
Given an image classifier $\mathcal{N}$ and an input image $x$, we define a
binary causal model $M_{\mathcal{N},x}$ as follows. The set $\V = \vec{V} \cup \{O\}$ of endogenous variables consists
of a set $\vec{V}$ corresponding to the set of pixels $P(x)$ of $x$ and the single output variable $O$. 
Essentially, $\vec{V}$ is a binary \emph{mask}, indicating which pixels of $x$ are visible and which are occluded, and
the output variable $O$ indicates whether the classification of a partially masked image stays the same as of the original image.

Assigning $1$ to a variable $v_i \in \vec{V}$ means that the pixel $p_i$, corresponding
to $v_i$, has its original value (taken from $x$). Assigning $0$ to this variable means that $p_i$ is masked --
replaced with some predefined masking value.

The masking operation of $\V$ on $P(x)$ is denoted by $\V \odot P(x)$ and
is the Hadamard product of these sets viewed as matrices of the same size (corresponding to the input size and shape of $\mathcal{N}$).

The context $\vec{u}$ that assigns all variables in $\vec{V}$ the value $1$ (i.e., 
none of the pixels are masked) corresponds to the fully unmasked image $x$. 
The value of $O$ is $1$ iff the output of $\mathcal{N}$ on $\V \odot P(x)$,
the partially masked image defined by applying $\vec{V}$ to $x$, is $\mathcal{N}(x)$ and is $0$ otherwise. 
Clearly, $O=1$ if the image is fully unmasked, that is, $(M_{\mathcal{N},x},\vec{u}) \models (O=1)$.
We depict the structure of $M_{\mathcal{N},x}$ in \Cref{fig:MNx}.
The causal model has depth $2$. In what follows, we omit the subscript ${\mathcal{N},x}$ from the causal model notation
if it is clear from the context.

\begin{figure*}[t]
    \centering
    \begin{tikzpicture}[outer sep=auto]
    \node (V) at (-1, 0) {$\vec{V}$};
    \node (v1) at (0, 0) [draw, circle, minimum size=0.9cm,fill=red!10] {$v_1$};
    \node (v2) at (1.5, 0) [draw, circle, minimum size=0.9cm,fill=red!10] {$v_2$};
    \node (v3) at (3, 0) [draw, circle, minimum size=0.9cm,fill=red!10] {$v_3$};
    \node (v4) at (4.5, 0) [draw, circle, minimum size=0.9cm,fill=red!10] {$v_4$};
    \node (v5) at (6, 0) [draw, circle, minimum size=0.9cm,fill=red!10] {$v_5$};
    \node (dots) at (7, 0) {$\mathbf{\cdots}$};
    \node (v6) at (8, 0) [draw, circle, minimum size=0.3cm,fill=red!10] {$v_{n-2}$};
    \node (v7) at (9.5, 0) [draw, circle, minimum size=0.3cm,fill=red!10] {$v_{n-1}$};
    \node (vn) at (11, 0) [draw, circle, minimum size=0.9cm,fill=red!10] {$v_n$};

    \node (f) at (5.5, -2.2) [draw, rectangle, rounded corners, fill=blue!10, inner sep=0.25cm] 
    {$\mathcal{N}(\vec{v} \odot P(x)) \? \mathcal{N}(x)$};

    \draw [-Triangle] (v1) -- (f);
    \draw [-Triangle] (v2) -- (f);
    \draw [-Triangle] (v3) -- (f);
    \draw [-Triangle] (v4) -- (f);
    \draw [-Triangle] (v5) -- (f);
    \draw [-Triangle] (v6) -- (f);
    \draw [-Triangle] (v7) -- (f);
    \draw [-Triangle] (vn) -- (f);
    
    \node (o) at (5.5, -3.8) [draw, circle, minimum size=0.9cm, fill=yellow!10] {$O$};
    \node (oin) at (7, -3.8) {$O \in \{0,1\}$};

    \draw [-Triangle] (f) -- (o);
    \end{tikzpicture}
    \caption{A depth-$2$ binary causal model $M_{\mathcal{N},x}$ for an image $x$ and a classifier $\mathcal{N}$. $\vec{v}$ is the vector
    of values of $\vec{V}$. The output $O \in \{0, 1\}$ indicates
    whether the classification of the Hadamard product of the matrix of pixels of $x$ and $\vec{v}$ is the same as the original classification.}
    \label{fig:MNx}
\end{figure*}

\paragraph*{Causal Independence:}~\Cref{fig:MNx} assumes causal independence between the variables in $\vec{V}$. 
This is common to many approaches in causal and counterfactual explainable AI~\citep{USL19,SHG20,PSSBF20,MMTS21,Beckers22}, and is the \emph{de facto} approach in all black-box \xai tools. The reader is referred to~\citep{CH24,CKKS24} for an extended discussion on causal independence and
viewing images as data.

\commentout{
We argue, however, that this is a sound assumption in the case
of images, and not just a convenient approximation. Image classification models perform classifications over data, not over what the image represents. This data encodes correlations which are due to the underlying data production method being causal, \ie the real world. Obscuring or permuting pixels does not lead to a cascading change of other pixels values, as one would expect if pixels were causally connected, instead these permutations disrupt only correlations in the data. Pixels are not concepts: a $2$D image is a projection of a scene in the $3$D real world onto a plane; concepts that are present in the object in a real world can be invisible on this projection, hence the pixel independence.
Moreover, for each setting $\vec{v}$ of the feature
variables, there is a setting of the exogenous variables such that $\vec{V} = \vec{v}$.
That is, the variables in $\vec{V}$ are causally independent and
determined by the context.  Moreover, all the parents
of the output variable $O$ are contained in $\vec{V}$.  
Given these assumptions, the probability on contexts directly
corresponds to the probability on seeing various images which the neural network
presumably learns during training.
}

%proved in~\cite{CH24} to be equivalent to~\Cref{def:EX}
%(their proof is for a partial explanation, which is a generalization of the notion of explanation).
Given a neural network $\mathcal{N}$ and an input image $x$, let $\vec{u}_1$ be the context
that assigns $1$ to all variables in $\vec{V}$, and let $\vec{u}_0$ be the context
that assigns $0$ to all these variables, \citet{CKKS24} introduce the following definition. 
%We introduce the following definition of explanation for image classifiers.

\dfn[Single-Context Sufficient Explanation~\citep{CKKS24}]\label{defn:simple-exp}
A subset $\vec{V}_{exp}$ of $\vec{V}$ is a \emph{single-context sufficient explanation} of a classification $\mathcal{N}(x)$ of 
an input image $x$ by a classifier 
$\mathcal{N}$ if the following conditions hold:
\begin{description}
\item[{\rm EXIM1.}] $(M,\vec{u}_0) \models [\vec{V}_{exp} = 1](O=1)$.
\item[{\rm EXIM2.}] $\vec{V}$ is minimal; there is no strict subset $\vec{V}'$ of $\vec{V}_{exp}$ that satisfies EXIM1,
where $\vec{v}'$ is the restriction of $\vec{v}$ to the variables in $\vec{V}'$.
\end{description}
As there is a one-to-one correspondence between the variables in $\vec{V}$ and the pixels of $x$, 
we also call the subset of pixels $P_{exp}$ of $x$ that corresponds to $\vec{V}_{exp}$ an \emph{explanation} of $\mathcal{N}(x)$.
In other words, it is a minimal subset of pixels of a given input image $x$ that is sufficient for the 
model $\mathcal{N}$ to classify the image, with all other pixels masked.
Note that we do not assume that the classification of a fully masked input by $\mathcal{N}$ is different from $\mathcal{N}(x)$; if they are equal, the (single) explanation is an empty set.
\edfn
\citet{CKKS24} prove that in the context of image classifiers, \Cref{defn:simple-exp} is equivalent to \Cref{def:AC} in the context
that assigns $0$ to all variables in $\vec{V}$.

\subsection{Logic-based Explanations}\label{subsec:logic}

% \begin{figure*}
%     \centering
%     \begin{subfigure}{0.24\textwidth}
%         \centering
%         \includegraphics[viewport=80 100 180 150,clip]{monotonicity/01_301.png}
%         \caption{301: ladybug}%
%         \label{fig:lady_19}
%     \end{subfigure}
%     \hfill
%     \begin{subfigure}{0.24\textwidth}
%         \centering
%         \includegraphics[viewport=80 100 180 150,clip]{monotonicity/02_301.png}
%         \caption{301: ladybug}%
%         \label{fig:lady_20}
%     \end{subfigure}
%     \hfill
%     \begin{subfigure}{0.24\textwidth}
%         \centering
%         \includegraphics[viewport=80 100 180 150,clip]{monotonicity/03_27.png}
%         \caption{27: adolescent newt (eft)}%
%         \label{fig:lady_21}
%     \end{subfigure}
%     \hfill
%      \begin{subfigure}{0.24\textwidth}
%         \centering
%         \includegraphics[viewport=80 100 180 150,clip]{monotonicity/04_301.png}
%         \caption{301: ladybug}%
%         \label{fig:lady_22}
%     \end{subfigure}
%     \caption{Non-monotonicity in image classifiers. Each image is obtained from the previous one by adding $10$ pixels only, but the classification
%     changes in~\Cref{fig:lady_21} from `ladybug' to `adolescent newt (eft)', and then back to `ladybug' in~\Cref{fig:lady_22}. The explanations
%     are the highlighted pixels, and the rest of the image, shown as dimmed, is masked out.}
%     %\hfill
% \end{figure*}

We now briefly review some relevant definitions from the world of logic-based explanations. 

A classification problem is characterized by a set of features $\mathcal{F} = \{1 \dots m\}$ and a set of classes $K = \{c_1, \dots c_k\}$.
Each feature $i \in \mathcal{F}$ has a domain $D_i$, resulting in a feature space $\mathbb{F} = D_1 \times D_2  \dots \times D_m$. 
The classifier $\mathcal{N}$ cannot be a constant function: there must be at least two different points in the feature space that have different classifications. 
The most important assumption underlying the computability of logic-based explanations is monotonicity.

\dfn\label{def:monotonicity}[Monotonic Classifier~\citep{marques21a}]
Given feature domains and a set of classes assumed to be totally ordered, 
a classifier $\mathcal{N}$ is fully monotonic if $a \leq b \Rightarrow \mathcal{N}(a) \leq \mathcal{N}(b)$ (where, given
two feature vectors $a$ and $b$, we say that $a \leq b$ if $a_i \leq b_i (i = 1, \dots n))$.
\edfn

\dfn\label{def:pi}[Abductive Explanation~\citep{marques21a}]
An \textit{abductive}, or \textit{prime-implicant} (PI), explanation is a subset-minimal set of features $\mathcal{X} \subseteq \mathbb{F}$, which, if assigned the values $v$ dictated by the instance $(v, c)$, are sufficient for the prediction $c$.
\begin{equation}\label{eq:abductive}
    \forall (x \in \mathbb{F}) \big[\bigwedge_{i \in \mathcal{X}} (x_i = v_i)\big] \rightarrow (\mathcal{N}(x) = c).
\end{equation}
\edfn
The notion of the relevant set of contexts is implicit in this definition. If we are to spell it explicitly, we say that
\Cref{eq:abductive} is satisfied on all contexts in the feature space $\mathbb{F}$.

The other common definition in logic-based explanations relevant to our discussion is \emph{contrastive} explanations. A contrastive explanation answers the question ``why did this happen, and not that?''~\citep{Mil19}.

\dfn\label{def:contrastive}[Contrastive Explanation~\citep{INAMS20}]
A \textit{contrastive} explanation is a subset-minimal set $\mathcal{Y} \subseteq \mathcal{F}$ which, if the features in 
$\mathbb{F} \setminus \mathcal{Y}$ are assigned the values dictated by the instance $(v, c)$ then there is an assignment to the features in $\mathcal{Y}$ that changes the prediction.  
\begin{equation}\label{eq:contrastive}
    \exists (x \in \mathbb{F}) \big[ \bigwedge_{i \in \mathcal{F} \setminus \mathcal{Y}} (x_i = v_i) \big] \land (\mathcal{N} (x) \not= c).
\end{equation}
\edfn

\section{Definitions}\label{sec:def}

In this section we suggest a set of definitions for explanations. The overall structure is similar to \Cref{defn:simple-exp}, with the
``moving parts'' being sufficiency vs necessity, single context vs multiple contexts, and confidence of the network in the result. All these
definitions are with respect to a given classifier $\mathcal{N}$ on a given image $x$ and the depth-$2$ causal model $M$ that is
constructed as described in \Cref{subsec:cause-img}. As usual, while we define subsets of $\vec{V}$ as explanations, the matching subsets of
pixels of $x$ are also called explanations in the same way.

For the sufficiency condition we are examining the result of
setting $\vec{V}_{exp}$ to $1$ on the context $\vec{u}_0$, that is, the whole image except $\vec{V}_{exp}$ is masked, and the pixels
corresponding to $\vec{V}_{exp}$ retain their original values. For the necessity condition, $\vec{V}_{exp}$ is masked, and the rest of the image
retains its original values. The original definition of an explanation in image classification, \Cref{defn:simple-exp}, is for
a \emph{single-context sufficient explanation}, and in what follows, we abbreviate it as \emph{SCSE}. The definition of a
\emph{(single-context) necessary explanation}, abbreviated to \emph{NE}, is obtained from \Cref{defn:simple-exp} by replacing EXIM1 
(and the reference to EXIM1 in EXIM2) with
\[ \text{EXN1. } (M,\vec{u}_1) \models [\vec{V}_{exp} = 0](O=0).\]

We further extend the causal framework in~\citep{CKKS24} by generalizing the concept of the masking value, which they take to be a  
a single pre-defined value that is the same in all masking operations. 
In this paper, we introduce a masking function $g: P \rightarrow D$, where $D$ is the range of values for pixels, 
representing their color and intensity. The introduction of
a function instead of a single value allows, for example, to use another image for masking, effectively resulting in overlaying a
candidate explanation on top of another image.

\Cref{defn:simple-exp} assumes a single masking value and a single masking operation that masks all pixels not in the candidate explanation.
The first natural step would be to extend EXIM1 to the set of contexts representing all partial maskings of the input image. The updated
condition is as follows, where $\vec{u}_{\vec{W}}$ is a context that assigns $1$ to a subset $\vec{W}$ of $\vec{V}$ and 
$0$ to $\vec{V} \setminus \vec{W}$.
\[ \text{EXIM1'. } \forall \vec{W} \subseteq \vec{V}. (M,\vec{u}_{\vec{W}}) \models [\vec{V}_{exp} = 1](O=1). \]
It might seem that EXIM1 implies EXIM1'. Indeed, if unmasking just $\vec{V}_{exp}$ already results in the original classification,
then unmasking more of the image would not change it. A classifier that has this property is \emph{monotonic}. 
However, interestingly, image classifiers are \emph{not monotonic}, as we show in~\Cref{fig:ladybird}: ~\Cref{subfig:lady_suff} is 
an SCSE for the classification ``ladybug''; adding just $20$ more pixels from the original image (\Cref{subfig:lady_21}) 
changes the classification to ``adolescent newt (eft)''. The whole image is classified as ``ladybug''.

To formally extend \Cref{defn:simple-exp} to multiple contexts, we introduce a set of
contexts $\K$, which are the result of applying different masking functions to $\vec{u}_1$. The definition of a \emph{multi-context
sufficient explanation (MCSE)} is obtained from \Cref{defn:simple-exp} by replacing EXIM1 and a reference to it in EXIM2 with
\[ \text{EXMC1. } \forall \vec{u} \in \K, (M,\vec{u}) \models [\vec{V}_{exp} = 1](O=1).\]
We note that there is no natural extension of necessary explanations to multiple contexts.

For the rest of the discussion in this paper, it is useful to introduce a stronger notion of explanation, namely \emph{complete explanations}.
Completeness is mentioned in work on saliency methods~\citep{srinivas19}, but is used there in a different meaning. 
In saliency methods, the intuition is, if the saliency map $S(x)$ completely encodes the computational information as performed by $\mathcal{N}$, 
then it is possible to recover $\mathcal{N}(x)$ by using $S(x)$ and $x$ using some function $\phi$. 
In effect, this means that, in addition to recovering the original model decision, we should also be able to recover the model's confidence in its decision.
As a causal explanation is not a saliency map, but rather a set of pixels, we cannot use this property directly.
%Informally, \textbf{a \onecomplete explanation for an image $x$ is a subset-minimal set of pixels which have both the same class and
%confidence as $\mathcal{N}(x)$ for some model $\mathcal{N}$.}
Informally, a complete explanation is an explanation that is both sufficient and necessary.
\dfn[Complete Explanation\label{def:complete}]
A subset $\vec{V}_{exp}$ of $\vec{V}$ is a \emph{single-context complete explanation (SCCE)} of a classification $\mathcal{N}(x)$ of 
an input image $x$ by a classifier $\mathcal{N}$ if it satisfies EXIM1 and EXN1, and there is no strict subset of $\vec{V}_{exp}$
that satisfies both EXIM1 and EXN1.

Similarly, $\vec{V}_{exp}$ is a \emph{multi-context complete explanation (MCCE)} if satisfies EXMC1 and EXN1
(note that necessary explanations are always single-context), and there is no strict subset of $\vec{V}_{exp}$
that satisfies both EXMC1 and EXN1.
\edfn

Finally, we discuss the confidence of the model $\mathcal{N}$. Recall, that the output of $M_{\mathcal{N},x}$ in \Cref{fig:MNx}
only depends on whether $\mathcal{N}(\vec{v} \odot P(x))$ is equal to $\mathcal{N}(x)$, without taking the confidence of $\mathcal{N}(x)$ into
account. As our experiments show, allowing a significantly lower confidence for an explanation than for the original image might result
in low-quality explanations (see the supplementary material). We, therefore, introduce \emph{$\delta$-confident} explanations. 
\dfn[$\delta$-confident Explanation\label{def:confident}]
An explanation $\vec{V}_{exp}$ is \emph{$\delta$-confident} if the confidence of $\mathcal{N}$ $(\vec{v}' \odot P(x))$ is at least
$\delta\cdot c$, where $c$ is the confidence of $\mathcal{N}(x)$, and $\vec{v}'$ is the mask computed from $\vec{V}_{exp}$ according to
whether it is a sufficient or a necessary explanation. If $\vec{V}_{exp}$ is sufficient, then $\vec{v}'$ is $1$'s for all 
$V \in \vec{V}_{exp}$ and $0$ for the other variables, and if $\vec{V}_{exp}$ is necessary, then $\vec{v}'$ is $0$'s for all 
$V \in \vec{V}_{exp}$ and $1$ for the other variables.
If the confidence of the classification of $\vec{V}_{exp}$ is exactly the same as the one of the original image, we say that
$\vec{V}_{exp}$ is \emph{$1$-exact confident} (shortened to \emph{$1$-confident} when there is no ambiguity).
A complete explanation is $\delta$-confident if it is $\delta$-confident both as a sufficient and as a necessary explanation.
\edfn
For brevity, we call explanations $\delta$-sufficient and \deltacomplete instead of $\delta$-confident sufficient and $\delta$-confident complete,
respectively.

As we discuss in more detail in \Cref{sec:algo}, \onecomplete explanations are of a special interest, as they are, in essence, equivalent to
the whole image from the point of view of the classifier. As we show, pixels that are not a part of either sufficient or necessary explanation,
but are needed to adjust the confidence of the classifier, can provide interesting insights into the decision process of the classifier. We
call these pixels \emph{adjustment} pixels.

\Cref{fig:exp} is a graphic depiction of the explanation landscape presented in this section.

\begin{figure}[t]
\centering
\begin{tikzpicture}
    \pgfmathsetmacro{\cubex}{1.5}
    \pgfmathsetmacro{\cubey}{1.5}
    \pgfmathsetmacro{\cubez}{1.5}

    % left bottom back -- empty
    %\draw[red!20] (1,1.2,0) -- ++(-\cubex,0,0) -- ++(0,-\cubey,0) -- ++(\cubex,0,0) -- cycle;
    %\draw[red!20] (1,1.2,0) -- ++(0,0,-\cubez) -- ++(0,-\cubey,0) -- ++(0,0,\cubez) -- cycle;
    %\draw[red!20] (1,1.2,0) -- ++(-\cubex,0,0) -- ++(0,0,-\cubez) -- ++(\cubex,0,0) -- cycle;

    % right botton back -- empty
    %\draw[red!20] (3.7,1.2,0) -- ++(-\cubex,0,0) -- ++(0,-\cubey,0) -- ++(\cubex,0,0) -- cycle;
    %\draw[red!20] (3.7,1.2,0) -- ++(0,0,-\cubez) -- ++(0,-\cubey,0) -- ++(0,0,\cubez) -- cycle;
    %\draw[red!20] (3.7,1.2,0) -- ++(-\cubex,0,0) -- ++(0,0,-\cubez) -- ++(\cubex,0,0) -- cycle;
    
    % right bottom front -- empty
    %\draw[red,fill=white] (2.7,0,0) -- ++(-\cubex,0,0) -- ++(0,-\cubey,0) -- ++(\cubex,0,0) -- cycle;
    %\draw[red,fill=white] (2.7,0,0) -- ++(0,0,-\cubez) -- ++(0,-\cubey,0) -- ++(0,0,\cubez) -- cycle;
    %\draw[red!30,fill=white] (2.7,0,0) -- ++(-\cubex,0,0) -- ++(0,0,-\cubez) -- ++(\cubex,0,0) -- cycle;
    
    % NE
    \draw[red,fill=yellow!10] (0,0,0) -- ++(-\cubex,0,0) -- ++(0,-\cubey,0) -- ++(\cubex,0,0) -- cycle;
    \draw[red,fill=yellow!10] (0,0,0) -- ++(0,0,-\cubez) -- ++(0,-\cubey,0) -- ++(0,0,\cubez) -- cycle;
    \draw[red!30,fill=yellow!10] (0,0,0) -- ++(-\cubex,0,0) -- ++(0,0,-\cubez) -- ++(\cubex,0,0) -- cycle;
    \node (ne) at (-.76, -.75) {NE};

    % delta-confident SCSE
    \draw[red!30,fill=red!10,opacity=0.5] (1,2.8,0) -- ++(-\cubex,0,0) -- ++(0,-\cubey,0) -- ++(\cubex,0,0) -- cycle;
    \draw[red!30,fill=red!10,opacity=0.5] (1,2.8,0) -- ++(0,0,-\cubez) -- ++(0,-\cubey,0) -- ++(0,0,\cubez) -- cycle;
    \draw[red!30,fill=red!10,opacity=0.5] (1,2.8,0) -- ++(-\cubex,0,0) -- ++(0,0,-\cubez) -- ++(\cubex,0,0) -- cycle;
    \node[align=center,text=black!80] (dscse) at (0.25, 2.3) {\small{$\delta$-confident}\\SCSE};
    
    % SCSE
    \draw[red,fill=orange!30,opacity=0.4] (0,1.6,0) -- ++(-\cubex,0,0) -- ++(0,-\cubey,0) -- ++(\cubex,0,0) -- cycle;
    \draw[red,fill=orange!30,opacity=0.4] (0,1.6,0) -- ++(0,0,-\cubez) -- ++(0,-\cubey,0) -- ++(0,0,\cubez) -- cycle;
    \draw[red,fill=orange!30,opacity=0.4] (0,1.6,0) -- ++(-\cubex,0,0) -- ++(0,0,-\cubez) -- ++(\cubex,0,0) -- cycle;
    \node (scse) at (-.76, 1.1) {SCSE};

    % delta-confident MCSE
    \draw[red!30,fill=red!10,opacity=0.5] (3.7,2.8,0) -- ++(-\cubex,0,0) -- ++(0,-\cubey,0) -- ++(\cubex,0,0) -- cycle;
    \draw[red!30,fill=red!10,opacity=0.5] (3.7,2.8,0) -- ++(0,0,-\cubez) -- ++(0,-\cubey,0) -- ++(0,0,\cubez) -- cycle;
    \draw[red!30,fill=red!10,opacity=0.5] (3.7,2.8,0) -- ++(-\cubex,0,0) -- ++(0,0,-\cubez) -- ++(\cubex,0,0) -- cycle;
    \node[align=center,text=black!80] (dmcse) at (2.95, 2.3) {\small{$\delta$-confident}\\MCSE};

     % MCSE
    \draw[red,fill=orange!30,opacity=0.4] (2.7,1.6,0) -- ++(-\cubex,0,0) -- ++(0,-\cubey,0) -- ++(\cubex,0,0) -- cycle;
    \draw[red,fill=orange!30,opacity=0.4] (2.7,1.6,0) -- ++(0,0,-\cubez) -- ++(0,-\cubey,0) -- ++(0,0,\cubez) -- cycle;
    \draw[red,fill=orange!30,opacity=0.4] (2.7,1.6,0) -- ++(-\cubex,0,0) -- ++(0,0,-\cubez) -- ++(\cubex,0,0) -- cycle;
    \node (mcse) at (2, 1.1) {MCSE};

    \node (single) at (-0.5, -2) {Single};
    \node (multiple) at (2,-2) {Multiple};
    \draw (single) edge[->] (multiple);

    \node (necessary) at (-2.5, -1) {Necessary};
    \node (sufficient) at (-2.5, 1) {Sufficient};
    \draw (necessary) edge[->] (sufficient);

    \node (nod) at (3.6, -1.3) {No $\delta$};
    \node (d) at (4.5, -0.3) {$\delta$};
    \draw (nod) edge[->] (d);

    \node[text=blue] (scce) at (-.75, 0) {SCCE};
    \draw[draw=blue,fill=blue] (ne) edge[-Triangle] (scce);
    \draw[draw=blue,fill=blue] (scse) edge[-Triangle] (scce);

    \node[text=blue] (mcce) at (2.4, -1.5) {MCCE};
    \draw[draw=blue,fill=blue] (ne) edge[-Triangle] (mcce);
    \draw[draw=blue,fill=blue, bend left] (mcse) edge[-Triangle] (mcce);

    \node[align=center,text=blue!60] (dscce) at (0.6, 0.6) {\small{$\delta$-confident}\\SCCE};
    \draw[draw=blue!80,fill=blue!50] (ne) edge[-Triangle] (dscce);
    \draw[draw=blue!50,fill=blue] (dscse) edge[-Triangle] (dscce);

    \node[align=center,text=blue!60] (dmcce) at (1.7, -0.5) {\small{$\delta$-confident}\\MCCE};
    \draw[draw=blue!80,fill=blue!50] (ne) edge[-Triangle] (dmcce);
    \draw[draw=blue!50,fill=blue!50] (dmcse) edge[-Triangle] (dmcce);

% minimum width = 2cm, 
% 	minimum height = 1.2cm
    %\node[ellipse, draw=brown, text=orange, minimum height=2.5cm] (e) at (-0.75,0) {SCCE};
    
\end{tikzpicture}
\caption{The summary figure of explanation types}%
\label{fig:exp}
\end{figure}

\section{Theoretical Results}\label{sec:theory}

In this section we prove our main theoretical results. 

\subsection{Results on our definitions} 
We start with a number of results for our definitions in~\Cref{sec:def} (see the supplementary material for the proofs).

\begin{lemma}\label{lemma-MSCE}
    MSCE is equivalent to \Cref{def:EX} in our setting.
\end{lemma}

A number of lemmas below are based on the following result.
\begin{lemma}\label{lemma-subset}
    For every MCSE $\vec{V}_1$ over the set of contexts $\K$ such that $\vec{u}_0 \in \K$, 
    there exists an SCSE $\vec{V}_2$ such that $\vec{V}_1 \subseteq \vec{V}_2$. 
\end{lemma}

\begin{lemma}\label{lemma-suf-nec}
    For monotonic classifiers, every sufficient explanation intersects with every necessary explanation.
    For all classifiers, every MCSE intersects with every necessary explanation.
\end{lemma}

\begin{corollary}\label{cor-complete}
For monotonic classifiers, any two complete explanations have a non-empty intersection. 
The result holds also for $\delta$-complete explanations, for any $0 \leq \delta \leq 1$. For all classifiers, the result holds for MCCE and for 
$\delta$-MCCE.
\end{corollary}

We note that it is straightforward to show that sufficient and necessary explanations exist. Indeed, the maximal size of such an explanation
is the whole $\vec{V}$, corresponding to the whole image. This also implies the existence of complete and $\delta$-complete explanations,
for all $0 \leq \delta \leq 1$. For $\delta > 1$, $\delta$-confident explanations might not exist: 
consider, for example, an input $x$ that is classified by $\mathcal{N}$ with confidence $1$; as the maximal confidence is $1$, no 
$\delta$-confident explanation for $\mathcal{N}(x)$ exists, for $\delta > 1$.

%\begin{lemma}\label{lemma:exist}
%    For any model $\mathcal{N}$ and any input image $x$, there always exist causal explanations for $\mathcal{N}(x)$ -- sufficient, necessary,
%    and complete. $\delta$-confident explanations exist for all $0 \leq \delta \leq 1$. 
%\end{lemma}

\subsection{Causal vs logic-based explanations}
We now turn to the formalization of logic-based explanations in the actual causality framework.

For a given classification problem as defined in \Cref{subsec:logic}, we define a depth-$2$ causal model $M$ as follows.
The set of endogenous input variables is the set of features $\mathcal{F}$, with each variable $i \in \mathcal{F}$
having a domain $D_i$. The output of the classifier is the output variable $O$ of the model, with the domain
$K$. An instance $(v,c)$ corresponds to a context $\vec{u}$ for $M$ that assigns to $\mathcal{F}$ the values defined by $v$,
and the output is $c$. The set $\mathcal{K}$ of contexts is defined as the feature space 
$\mathbb{F} = D_1 \times D_2  \dots \times D_m$. As the classifier is not constant, there exist at least two inputs
$v$ and $v'$ such that $(M,v) \sat O=c$, $(M,v') \sat O=c'$, with $c \neq c'$. It is easy to see that $M$ is a depth-$2$
causal model with all input variables being causally independent.
%\subsection{Equivalences between definitions}
Armed with this translation, we prove a number of equivalences between the causal and the logic-based explanations.

\begin{lemma}\label{lemma:ac_abductive}
   Multiple-context sufficient explanations (MCSEs) on the set $\K$ of all possible contexts are equivalent to abductive explanations (\Cref{def:pi}).
\end{lemma}

The following is an easy corollary from \Cref{lemma:ac_abductive} when we observe that the proof does not use any unique
characteristics of image classifiers.
\begin{corollary}
\Cref{lemma:ac_abductive} holds for any binary depth-$2$ causal models.
\end{corollary}

\begin{lemma}\label{lemma:ac_contrastive}
   Contrastive explanations (\Cref{def:contrastive}) are equivalent to SCSEs in the same setting (and both are equivalent to actual causes in ~\Cref{def:AC}).
\end{lemma}

Let us now discuss a useful property of causal explanations, namely, \emph{input invariance}.
Input invariance~\citep{kindermans22invariance} is a property first defined for saliency-based methods. 
It is a stronger form of the property introduced by~\citet{srinivas19} as \emph{weak dependence}. 
Given two models, $\mathcal{N}_1(x)$ and $\mathcal{N}_2(x)$, which are identical other than the  $\mathcal{N}_2$ 
has had its first neuron layer before non-linearity altered in a manner that does not affect the gradient (\eg means-shifting)
from $\mathcal{N}_1(x)$, there should be no difference in saliency map, \ie $S(\mathcal{N}_1(x)) = S(\mathcal{N}_2(x))$. 

Some methods, such as LRP~\citep{bach2015pixel}, do not satisfy input invariance~\citep{kindermans22invariance}. 
Other methods, notably LIME~\citep{lime} train an additional model on local data perturbations: 
it is difficult to make a general statement regarding LIME and input invariance due to local model variability.

As a causal explanation is independent of the exact values of $x$, and depends only on the output of $\mathcal{N}$, a causal explanation is invariant in the face of such alterations. The only subtlety being that the masking function, $g$, also needs to be means-shifted.
The following lemma follows from the observation that \Cref{defn:simple-exp} depends only on the properties of $x$ and not on its values.
\begin{lemma}\label{lemma:input_invariant}
   All versions of causal explanations are input invariant.
\end{lemma}

\subsection{Complexity results}
As a motivation for the approximation algorithms described in \Cref{sec:algo}, we prove that all types of causal explanations are intractable.
\citet{CKKS24} proved that SCSEs are co-NP-complete. The following theorem fills the gap for other types of explanations. We note that the hardness of MCSE in co-NP follows from the hardness of SCSE, but the membership in the same complexity class is somewhat 
surprising, given that MCSE is defined over a set of contexts. 

\begin{theorem}\label{theor:complexity}
The decision problems of MCSE, NE, SCCE, and MCSE are co-NP-complete.    
\end{theorem}
The following result follows directly from \Cref{theor:complexity} and \Cref{lemma:ac_abductive,lemma:ac_contrastive}.
\begin{corollary}
    Abductive and contrastive explanations always exist. The decision problems of abductive and contrastive explanations are co-NP-complete.    
\end{corollary}

\section{Algorithm}\label{sec:algo}

\begin{figure*}[t]
    \centering
    \begin{subfigure}{0.23\textwidth}
        \centering
        \includegraphics[scale=0.2]{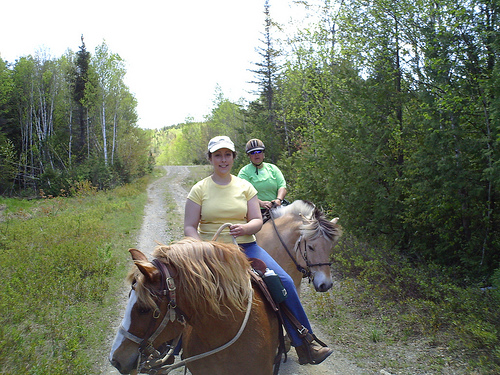}
        \caption{Original image, misclassified as ox ($0.2919$ confidence)}%
        \label{fig:ox}
    \end{subfigure}
    \hfill\hfill
    \begin{subfigure}{0.23\textwidth}
        \centering
        \includegraphics[scale=0.35]{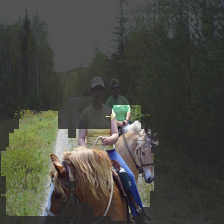}
        \caption{Sufficient explanation ($0.18$ confidence)}%
        \label{fig:ox_suff}
    \end{subfigure}
    \hfill
    \begin{subfigure}{0.23\textwidth}
        \centering
        \includegraphics[scale=0.35]{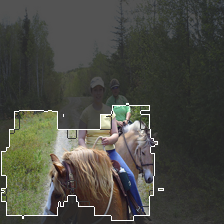}
        \caption{Contrastive (necessary) explanation $(0.298$ confidence)}%
        \label{fig:ox_con}
    \end{subfigure}
    \hfill
    \begin{subfigure}{0.23\textwidth}
        \centering
        \includegraphics[scale=0.35]{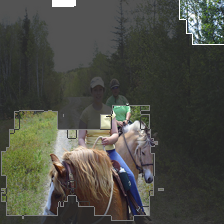}
        \caption{$1$-Complete explanation}%
        \label{fig:ox_comp}
    \end{subfigure}
    \caption{This image has the most extreme contrast classification for \mob{} on a \voc{} image. \mob{} incorrectly classifies this as \emph{ox} with a relatively low confidence (\Cref{fig:ox_suff}). The \deltacomplete explanation, which has a higher confidence than~\Cref{fig:ox} has the inverse classification of moped (confidence $0.133$). Finally, the adjustment pixels are classified as `picket fence'. The unusual behavior may be a result of the original misclassification.}
    \label{fig:inverse_example}
\end{figure*}

Given the intractability of all types of explanations as proved in~\Cref{theor:complexity}, we describe algorithms for
computing \emph{approximate explanations}. \Cref{algo:contrastive} presents the greedy approach for approximation of \deltacomplete 
multi-context ($\delta$-MCCE) explanations, with respect to the set of contexts consisting of all partial maskings of the input image.
In what follows, we omit ``multi-context'' for brevity.

We use \rex~\citep{CKKS24} as a basis for computing the definitions provided in this paper. 
\rex uses an approximation of causal responsibility to rank features.
Responsibility is a quantitative measure of causality, measuring, roughly speaking, the amount of causal influence a variables has on the classification~\citep{CH04}.
We use a \emph{responsibility map} created by \rex (\Cref{algo:contrastive},~\Cref{line:rank}) 
to rank pixels by their responsibility towards the desired classification. 

We use $2$ different sets of contexts, \kplus and \kminus (\Cref{algo:contrastive}, \Cref{line:contexts}) to approximate $\delta$-MCCEs.
\kplus is created by inserting pixels into an image created from the baseline defined by the masking function $g$, in the order of
their responsibility. \kminus does the opposite: it replaces pixels from $x$ with their masked values, also in the order
of their responsibility. 
The effect in practice is that at each step of the discovery procedure we consider two images, one the inverse of the other.
In the worst case, we need to explore all contexts in \kminus, as the entire image is required for completeness, 
however this worst case did not occur in our evaluation. 
% In our experiments, we use the same baseline value for both \kplus and \kminus. This is not a strict requirement of the algorithm or theory.

\begin{algorithm}[t]
  \caption{\deltacomplete Explanation $(x, \delta, \mathcal{N})$}
  \label{algo:contrastive}
  \begin{flushleft}
    \textbf{INPUT:}\,\, an image $x$, a confidence scalar $\delta$ in $(0, 1)$ and a model $\mathcal{N}$\\
    \textbf{OUTPUT:}\,\, a sufficient explanation $s$, a complete explanation $c$, a sorted responsibility ranking $\mathcal{R}$
  \end{flushleft}
  \begin{algorithmic}[1]
    \STATE $s, c \leftarrow $ initialize to $\emptyset$
    \STATE $l \leftarrow \mathcal{N}(x)$
    \STATE $\tau \leftarrow \delta \times \sigma(\mathcal{N}(x))$ (model confidence scaled by $\delta$)
    \STATE $\mathcal{R} \leftarrow \mathit{pixel\_ranking}(x, \mathcal{N})$ sorted \textbf{high} to \textbf{low}\label{line:rank}
    \STATE $\mathcal{K}^{+}, \mathcal{K}_{-} \leftarrow$ initialize \label{line:contexts}
    \FOR{$p \in \mathcal{R}$}
        \STATE $\mathcal{K}^{+}, \mathcal{K}_{-} \leftarrow$ update with $p$
         \IF{$\mathcal{N}(\mathcal{K}^+) = l$ and $\mathcal{N}(\mathcal{K}_{-}) \not= l$ and $\sigma(\mathcal{N}(\mathcal{K^+}))\geq \tau$}
             \STATE s $\leftarrow \mathcal{K}^+$
             \STATE c $\leftarrow \mathcal{K}_{-}$
             \STATE \textbf{return} $s, c, \mathcal{R}$
         \ENDIF
    \ENDFOR
  \end{algorithmic}
\end{algorithm}

~\Cref{algo:adjustment} details the procedure for discovering the adjustment pixels, required to turn a \deltacomplete into a \onecomplete explanation. %This is not always required, as a \deltacomplete explanation may already be \onecomplete.
In practice, we replace the exact equality of~\Cref{line:eq} with a user-provided degree of precision, set to $4$ decimal places by default.

\begin{algorithm}[t]
  \caption{{\onecomplete discovery}$(x, \mathcal{N}, \mathcal{R}, e$)}
  \label{algo:adjustment}
  \begin{flushleft}
    \textbf{INPUT:}\,\, an image $x$, a model $\mathcal{N}$, a responsibility landscape $\mathcal{R}$, a complete explanation $e$\\
    \textbf{OUTPUT:}\,\, a set of adjustment pixels $a$
  \end{flushleft}
  \begin{algorithmic}[1]
    \STATE $c \leftarrow \sigma(\mathcal{N}(x))$ (model confidence)
    \STATE $\mathit{pixel\_ranking} \leftarrow$ order\_pixels($\mathcal{R})$\label{line:order}
    \STATE $a \leftarrow \emptyset$
    \FOR{each pixel $p_i \in \mathit{pixel\_ranking}$}
         \STATE $a \leftarrow a \cup\{p_i\}$
         \STATE $e \leftarrow e \cup a$
         \STATE $x'\leftarrow$ mask pixels of $x$ that are \textbf{not} in e
         \IF{$\mathit{confidence}(\mathcal{N}(x')) = c $}\label{line:eq}
           \RETURN{$a$}
         \ENDIF
     \ENDFOR
  \end{algorithmic}
\end{algorithm}

The function `order\_pixels' (\Cref{algo:adjustment}, \Cref{line:order}) sorts the pixels by their responsibility in either ascending or descending order. The rationale behind this is, if the \deltacomplete explanation has a higher confidence than the original image, we add pixels with very low or $0$ responsibility in order to decrease the model's confidence. Conversely, if the \deltacomplete explanation has a confidence lower than the original score, we add pixels which have higher responsibility towards the classification. 
%Again, due to the non-monotonicity of image classifiers, we cannot easily take a more sophisticated search-based approach.

\begin{figure*}[t]
    \centering
    \scalebox{0.8}{\input{imagenet_results.pgf}}
    \caption{Results on \imagenet for $\delta = 1.0$. Both \swint and \resnet have very low requirements for sufficiency compared to \mob. \resnet also has the lowest requirement for $\delta$-completeness with confidence at least as great as the original image.}%
    \label{fig:results}
\end{figure*}

\begin{figure}[t]
    \centering
    \includegraphics[scale=0.4]{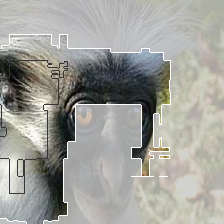}
    \caption{A \deltacomplete explanation with a low \imagenet distance. This colobus monkey has guenon monkey as its inverse classification. The model is clearly relying on the muzzle or snout to refine its classification to colobus.}
    \label{fig:monkey}
\end{figure}

\section{Experimental Results}\label{sec:eval}
\input{inverse}

In this section, we present an analysis of various models and datasets viewed through the lens of \deltacomplete and \onecomplete explanations. 
To the best of our knowledge, we are the first to compute \deltacomplete and \onecomplete explanations for image classifiers. Moreover, no-one has previously investigated the relationship between original and inverse classifications. 
Inverse classifications differ from simple counterfactuals  in that they 
characterise the data after \emph{all} relevant information have been removed. 
Due to the hierarchical nature of the \imagenet dataset, we can calculate the shortest path between the original classification and its inverse class. We do the same with the \onecomplete explanations, isolating the adjustment pixels from the \deltacomplete explanation, 
classifying them and calculating the shortest path to the original classification. 
For reasons of space, we include only a representative selection of results here. Complete results are presented in the supplementary material.

Our algorithms have been implemented as a part of the publicly available \xai tool \rex\footnote{\url{https://github.com/ReX-XAI/ReX}}. 
\rex was used with default parameters for all experiments, 
in particular, the random seed is $0$, the masking function $g$ is a constant function which sets all masked values to $0$, 
and the number of iterations is set to $20$.

Our experimental evaluation was performed on $3$ models, all from \textsc{TorchVision}: a \resnet, a \mob and \swint. 
All models were used with their default preprocessing procedures as 
provided by \textsc{TorchVision}. We applied these models to $3$ different standard and publicly available datasets: 
\imagenet-1K validation (approximately $4000$ images)~\citep{imagenet}, 
PascalVoc20212~\citep{pascal-voc-2012} and a dataset of complex images, ECSSD~\citep{ecssd}.
The experiments were done on NVidia A100 GPU running Ubuntu LTS 22.04. We 
found that runtime varied greatly depending on the model under test. The \resnet and \mob models were both very efficient, 
taking $\approx 6$ seconds per image. 
The \swint model was slower, taking $\approx 16$ seconds per image. 

\Cref{fig:results} shows the relative sizes of sufficient, \onecomplete and adjustment pixel sets for the $3$ different models on \imagenet. 
In general, \resnet requires the fewest pixels for both sufficiency and $\delta$-completeness, and also has very few adjustment pixels. 
\mob and \swint appear to be much more similar in their behavior, though \swint has slightly larger complete explanations in general.

\Cref{fig:inverse_distance} shows the shortest path between the original classification and its inverse class, according to the \imagenet hierarchy. In general, across all models, the distance between the two classes is not large, with a maximum distance of $24$. This is not always the case, however:~\Cref{fig:ox} shows an example of a (mis-)classification where the inverse classification is `moped'. The adjustment pixels are classified as `picket fence'. It is worth noting that the initial confidence was already low on this image. At the other extreme, \Cref{fig:inverse_distance} reveals a few cases where the distance between the original classification and its inverse was very small. Manual inspection shows that these cases represent small classification shifts within a larger `umbrella' category.~\Cref{fig:monkey}, for example, shows that \resnet model 
required the highlighted pixels to refine the classification to \emph{colobus} monkey. 
Without them, the classification is still monkey, but a different subclass -- guenon. 

\paragraph*{\xai Tool Comparison} In order to show the generality of our techniques, we applied our algorithms to the saliency output of two other popular \xai tools, \gradcam~\citep{selvaraju2017grad} and \lime~\citep{lime}. \gradcam is a white-box tool which accesses hidden layers in the model. \lime builds a locally interpretable model trained on datasets generated from the image to explain. While neither of these tools uses causality, they both produce output which can be used to rank pixels. We use this ranking as a surrogate for the causal responsibility ranking produced by \rex. 

We found that the ranking for all tools was good enough to discover sufficient explanations, \deltacomplete explanations and \onecomplete explanations. 
A natural comparison measure therefore, is to consider the precision of these explanations, \ie how many unnecessary pixels they contain. The more precise ranking should produce smaller explanations on average. 

We find that \rex finds the smallest sufficiencies in general. Across the $3$ different models, the average sufficiency size across all datasets for \rex was $2173.0$ pixels, or $\approx 4\%$ of the image. \lime performs well, with similar performance on sufficiencies to \rex, requiring on average $7036.5$ pixels, or $\approx 14\%$ of the image. \gradcam has very unpredictable behavior, needing only $
\approx 2561.1$ pixels on \resnet on average, but a massive $30464.6$ on \swint. It is worth noting that these sufficiencies were very close to being \deltacomplete however. This is not that surprising, as saliency patterns are architecture specific. Neither \rex nor \lime suffer from this architecture dependence problem. 

Interestingly, the different \xai methods differ most when calculating sufficiencies. Their patterns for $1$-completeness are very similar, \rex needing $\approx 53\%$ of the image, \gradcam $\approx 59\%$ and \lime $\approx 54\%$. \rex had the lowest average standard deviation, at $\pm 14246.5$, with \lime $\pm 14845.5$ and \gradcam $\pm 15484.5$. These results indicate that all the different ranking methods perform well, with \rex and \lime being more slightly stable than \gradcam. \lime uses a segmentation algorithm to partition the image under test. When this algorithm does not perform well~\citep{blake2024explainable}, \lime'{}s ranking suffers and this would also likely be reflected here. We leave this investigation to future work. \rex does not rely on any knowledge of the structure of the image contents which goes some way to explaining its reliability.

\paragraph{Related Work}
Broadly speaking, the field of \xai can be split between \emph{formal} and \emph{informal} methods~\citep{izza2024}. The majority of methods belong in the \emph{informal} camp, including well-known model agnostic methods~\citep{lundberg2017unified,lime} and saliency methods~\citep{CAM,bach2015pixel}. Formal explanation work has been dominated by logic-based methods~\citep{shih2018,ignatiev19}.
Logic-based explanations use abductive reasoning to find the simplest or most likely explanation `a' for a (set of) observations `b'. Logic-based explanations provide formal guarantees of feature sufficiency (\Cref{def:pi,def:contrastive}), but usually require strong assumptions of monotonicity or linearity for reasons of computability. This, together with weak scalability, poses questions about applicability of this approach to real settings.
Some logic-based methods are a \emph{black-box} \xai method, in that they do not require access to a model's internals, or even its gradient.

Constraint-driven black-box explanations~\citep{shrotri22} build on the LIME~\citep{lime} framework but include user-provided boolean constraints over the search space. In particular, for image explanations, these constraints could dictate the nature of the perturbations the \xai tool generates. Of course, knowing which constraints to use is a hard problem and assumes at least some knowledge of how the model works and what the explanation should be. While such methods are technically black-box, this is because model-dependent information has been separately encoded by the user.

Causal explanations~\citep{CH24} belong in the camp of formal \xai, as they provide mathematically rigorous guarantees in much the same manner as logic-based explanations.
\rex~\citep{CKKS24} is a black-box causal explainability tool which makes no assumptions about the classifier. It computes an approximation to minimal, sufficient pixels sets against a baseline value. 

We are not the first to offer definitions of sufficiency and necessity~\cite{bharti25a} for explanations, but ours use the language of actual causality. We also show that our definitions can be computed.

\section{Conclusions}\label{sec:conclusion}
 We introduced a set of definitions for causal explanations that cover sufficiency, necessity, completeness, and address confidence.
 We have demonstrated that logic-based explanations have natural equivalents in the actual causality framework.
 We also argued that by taking into account confidence, we can learn interesting properties of the model, in particular 
 by examining adjustment pixels---pixels that change the confidence of explanations. To the best of our knowledge,
 these aspects have not been studied before.
 
 We constructed and implemented algorithms for computing approximate explanations based on our definitions and merged them into 
 the existing open source causal
 explainability tool \rex. Our experimental results on $3$ standard datasets and $3$ standard models demonstrate significant differences between
 the models in the different types of explanations they induce, as well as in their behavior wrt confidence level.
 
 %We have created algorithms for approximating these definitions and incorporated them into the tool \rex. We have computed contrastive and complete 
 %explanations on $3$ different datasets with $3$ different models. We find that different models have different sufficiency, contrastive and 
 %adjustment requirements. Finally, we have examined the relationship between the original, contrastive and adjustment pixel predictions. 

\paragraph*{Acknowledgements}
The authors were supported in part by 
CHAI---the EPSRC Hub for Causality in Healthcare AI with Real Data (EP/Y028856/1).

%\newpage
\bibliographystyle{kr}
\bibliography{all}

\appendix

\section{Code and Data}
Our algorithms have been included in the open source \xai tool \rex. \rex is available at \url{https://github.com/ReX-XAI/ReX}. Complete data and analysis code can be found at the following anonymous link \url{https://figshare.com/s/7d822f952abcbe54ca93}.

\section{Proofs}
\setcounter{section}{4}

All numbered references match those in the main paper, unless explicitly stated otherwise.

\begin{lemma}\label{lemma-MSCE} 
MSCE is equivalent to Definition 2 in our setting.
\end{lemma}
\begin{proof}
  \citet{CH24} proved the following result (the proof in their paper is for
  partial explanations; we restate it for precise explanations here).
For the causal model $M$ corresponding to an image
  classifier and a set of contexts $\K$, $\vec{X} = \vec{x}$ is an explanation of $O=1$,
  if the following conditions hold:
  \begin{itemize}
  \item[\rm{EX1}] for all contexts $\vec{u} \in \K$, we have $(M,\vec{u}) \models [\vec{X} = \vec{x}](O=1)$;
  \item[\rm{EX2}] there is no strict subset $\vec{X}'$ of $\vec{X}$ such that 
      $(M,\vec{u}) \models [\vec{X}' = \vec{x}'](O=1)$,
      where $x'$ is the restriction of $\vec{x}$ to the variables in $X'$;
  \item[\rm{EX3}] there exists a context $\vec{u} \in \K$ and a setting $\vec{x}'$ to $\vec{X}$ such that
  $(M,\vec{u}) \models (\vec{X} = \vec{x}) \wedge (O = 1)$ and
  $(M,\vec{u}) \models [\vec{X} = \vec{x}'](O =0)$.
  \end{itemize}

The condition EX1 is the same as EXMC1. The condition EX2 is the minimality condition EXIM2. Finally, a context $\vec{u}$ in which EX3 holds is, for example,
the context that keeps all variables except those in $\vec{X}$ masked, and the variables in $\vec{X}$ have their original values.
\end{proof}

\begin{lemma}\label{lemma-subset}
    For every MCSE $\vec{V}_1$ over the set of contexts $\K$ such that $\vec{u}_0 \in \K$, 
    there exists an SCSE $\vec{V}_2$ such that $\vec{V}_1 \subseteq \vec{V}_2$. 
\end{lemma}
\begin{proof}
    We note that if $\vec{u}_0 \in \K$, then EXMC1 implies EXSC1. The minimality condition is stricter for MCSE, hence the MCSE is a subset (not necessarily strict)
    of the SCSE.
\end{proof}

\begin{lemma}\label{lemma-suf-nec}
For monotonic classifiers, every sufficient explanation intersects with every necessary explanation. 
    For all classifiers, every MCSE intersects with every necessary explanation.
\end{lemma}
\begin{proof}
    Recall that a set $\vec{V}_1$ is an MCSE if for every context $\vec{u} \in \K$, 
    $(M,\vec{u}_0) \models [\vec{V}_1 = 1](O=1)$.
    Similarly, $\vec{V}_2$ is a necessary explanation if $(M,\vec{u}_1) \models [\vec{V}_2 = 0](O=0)$.
    Assume the contrary, that is, $\vec{V}_1 \cap \vec{V}_2 = \emptyset$. Then, let $\vec{u}'$ be a context in which all variables in $\vec{V}_1$ are
    assigned $1$, and all variables in $\vec{V}_2$ are assigned $0$. Then, by the first equation, $O=1$, but by the second equation,
    $O=0$, which is a contradiction. Therefore, $\vec{V}_1 \cap \vec{V}_2 \not= \emptyset$.

    Now let $\vec{V}_1$ be an SCSE. If the classifier is monotonic, 
    then $(M,\vec{u}_0) \models [\vec{V}_1 = 1](O=1)$ implies that $(M,\vec{u}) \models [\vec{V}_1 = 1](O=1)$
    for all partial maskings of $\vec{V}$ (see the discussion in Section 4). Then, in particular, it holds for the context $\vec{u}'$ 
    defined as above, leading to a contradiction.

    We note that for non-monotonic classifiers, we can have non-intersecting SCSE and NE. This, in particular, would be the case with the example
    in Figure 1.    
\end{proof}

\begin{corollary}\label{cor-complete}
 Any two multi-context complete explanations have a non-empty intersection. The result holds also for $\delta$-complete explanations, for any 
 $0 \leq \delta \leq 1$. For monotonic classifiers, the result holds for single-context explanations as well.  
\end{corollary}
\begin{proof}
    The proof follows from the fact that a complete explanation is, in particular, a union of a sufficient and a necessary explanation. 
    Applying \Cref{lemma-suf-nec}, we get the result for MCCEs and for SCCEs in monotonic classifiers.
\end{proof}

\begin{lemma}\label{lemma:ac_abductive}
   Multiple-context sufficient explanations (MCSEs) on the set $\K$ of all possible contexts are equivalent to abductive explanations.
\end{lemma}
\begin{proof}
The definition of MSCE is has two conditions:
\[ \text{EXMC1. } \forall \vec{u} \in \K, (M,\vec{u}) \models [\vec{V}_{exp} = 1](O=1).\]
and the minimality condition EXIM2: there is no strict subset $\vec{V}'$ of $\vec{V}_{exp}$ that satisfies EXMC1. 
Equation 1 captures EXMC1, and EXIM2 is equivalent to the subset-minimality condition in the definition
of abductive explanations.
\end{proof}

The following is an easy corollary from \Cref{lemma:ac_abductive} when we observe that the proof does not use any unique
characteristics of image classifiers.
\begin{corollary}
\Cref{lemma:ac_abductive} holds for any binary depth-$2$ causal models.
\end{corollary}

\begin{lemma}\label{lemma:ac_contrastive}
   Contrastive explanations are equivalent to SCSEs in the same setting
   (and both are equivalent to actual causes).
\end{lemma}
\begin{proof}
Recall that the framework of logic-based explanations translates to causal models of depth $2$ with all 
input variables being independent. The condition AC1 in Definition 1 just states that in the current setting, $\vec{X}=\vec{x}$ and $O=o$. This is implicitly
assumed in the definition of contrastive explanations.
The condition AC2 is equivalent to Equation 2 if $\vec{W}=\emptyset$. We refer the reader to, in particular, \citep{CKKS24}, for the
statement that in depth-$2$ causal models, the condition AC2 is equivalent to the one with $\vec{W}=\emptyset$.
The condition AC3 is the minimality condition, stated in Definition 6 as subset-minimality.
The proof of equivalence of SCSEs and actual causes is similar. AC1 is implicit, as by construction 
$(M,\vec{u}) \models (\vec{V}_{exp} = 1) \wedge (O=1)$. AC2 is equivalent to EXIM1, and AC3 is the minimality condition equivalent to EXIM2.
\end{proof}

The following lemma follows from the observation that Definition 3 depends only on the properties of $x$ and not on its values.
\begin{lemma}\label{lemma:input_invariant}
   All versions of causal explanations are input invariant.
\end{lemma}

\begin{theorem}\label{theor:complexity}
The decision problems of MCSE, NE, SCCE, and MCSE are co-NP-complete.    
\end{theorem}
\begin{proof}
    \citet{CKKS24} prove co-NP-completeness of SCSE. It may seem that MCSE is a harder problem, as it generalizes the definition of SCSE to
    a set of contexts. In particular, co-NP-hardness of MCSE follows from co-NP-hardness of SCSE.

    It remains to prove the membership of MCSE in co-NP. We prove the membership of the complementary problem in NP. A subset $\vec{V}_1 \subseteq \vec{V}$
    is not an MCSE if either EXMC1 or the minimality condition are falsified. The falsification of condition EXMC1 means that there exists a context
    $\vec{u} \in \K$ such that $(M,\vec{u}) \not\models [\vec{V}_{exp} = 1](O=1)$, that is, $(M,\vec{u}) \models [\vec{V}_{exp} = 1](O=0)$.
    The falsification of the minimality condition means that there exists a subset $\vec{V}_2 \subset \vec{V}_1$ that satisfies EXMC1. 
    Given a witness -- either a context $\vec{u}$ or a subset $\vec{V}_2$ -- checking that the conditions are falsified is polynomial in the size of the input,
    hence the membership in NP. Therefore MCSE is in co-NP. Together with the NP-hardness result, it completes the proof for MCSE.

    The proof of co-NP-completeness of NE follows the similar lines. For the membership in co-NP, we prove that the complementary problem is in NP.
    Indeed, given a candidate subset $\vec{V}_1 \subset \vec{V}$, it is not a necessary explanation if it falsifies either EXN1 or the minimality condition.
    The falsification of EXN1 is verified in polynomial time, by setting all variables in $\vec{V}_1$ to $0$, and all other variables to $1$ and evaluating $O$.
    The falsification of the minimality condition is verified in polynomial time given a witness subset of $\vec{V}_1$, hence the membership of
    the complementary problem in NP, and thus the membership of NE in co-NP.

    For the hardness of NE in co-NP, we describe a reduction from SCSE to NE. Recall that 
    \[ \text{EXIM1. } (M,\vec{u}_0) \models [\vec{V}_{exp} = 1](O=1),\]
    and
    \[ \text{EXN1. } (M,\vec{u}_1) \models [\vec{V}_{exp} = 0](O=0).\]
    Given a depth-$2$ causal model $M_1$ with the output $O_1$ and a candidate subset $\vec{V}_{exp}$, 
    we construct a causal model $M_2$ with the output $O_2$ by inverting the values of
    all variables and inverting the outcome. That is, $O_2 = \neg{O_1}$. Then, the context $\vec{u}_0$ for the variables of $M_1$ is mapped
    into the context $\vec{u}_1$ for the variables of $M_1$, and hence 
    \[ (M_1,\vec{u}_0) \models [\vec{V}_{exp} = 1](O_1=1) \] 
    iff
    \[ (M_2,\vec{u}_1) \models [\vec{V}_{exp} = 0](O_2=0),\] 
    as required.

    The co-NP-hardness of complete explanations follows from the co-NP-hardness of sufficient and of necessary explanations. 
    The membership in co-NP follows from the observation that a witness to the falsification of the completeness conditions is verifiable in
    polynomial time.    
\end{proof}

\setcounter{section}{2}

\section{Experimental Results}

\begin{figure*}[t]
    \centering
    \scalebox{0.8}{\input{results.pgf}}
    \caption{Results on all $3$ datasets for $\delta=1.0$. Both \swint and \resnet have very low requirements for sufficiency compared to \mob. \resnet also has the lowest requirement for completeness with confidence at least as great as the original image.}%
    \label{fig:results}
\end{figure*}

\Cref{fig:results} shows the results across all $3$ datasets and models with $\delta = 1.0$. In general, all models follow a fairly similar pattern. It is interesting to note that \mob requires more pixels in general for a causal explanation, but also the lowest number of pixels for adjustment. This suggests that, for \mob at least, the \deltacomplete explanation encodes nearly all of its $1$-completeness into its \deltacomplete explanation.

\input{completeness}
\Cref{fig:adjustment_distance} shows the length of the shortest path between the original classification and the adjustment pixels.

\subsection{Different $\delta$}

\begin{figure*}[t]
    \centering
    \scalebox{0.8}{\input{results_05.pgf}}
    \caption{Results on all $3$ datasets for $\delta=0.5$. Both \swint and \resnet have very low requirements for sufficiency compared to \mob. \resnet also has the lowest requirement for completeness with confidence at least as great as the original image.}%
    \label{fig:results_05}
\end{figure*}

\Cref{fig:results_05} shows results for all datasets when the $\delta$-confident explanation threshold is $0.5$. 
If minimality is taken as a quality indicatory, this setting of $\delta$ sees a general deterioration of quality. The adjustment pixels sets are larger in general across all models and datasets. This suggests that this is a payoff between \deltacomplete and \onecomplete computationally: forcing the \deltacomplete explanation to have a higher confidence reduces the size of the adjustment pixel set, whereas a lower $\delta$ leads to a smaller \deltacomplete explanation and larger adjustment set. Users should bear this in mind when decided what aspect of a model's behavior they wish to explore.

\begin{figure*}[t]
    \begin{subfigure}[b]{0.3\textwidth}
        \centering
         \begin{tikzpicture}[scale=0.7]
            \begin{axis}[
            xlabel=Shortest Path (\imagenet-1K),
            ylabel=No. Images,
            xmin=0, xmax=24,
            ymin=0, ymax=800,
            ]
        \addplot[smooth,blue] plot coordinates {
            (2,212)
            (3,91)
            (4,250)
            (5,232)
            (6,285)
            (7,393)
            (8,471)
            (9,407)
            (10,350)
            (11,286)
            (12,253)
            (13,179)
            (14,130)
            (15,74)
            (16,35)
            (17,23)
            (18,5)
            (19,4)
        };
        \addlegendentry{\mob}
    
        \addplot[smooth,color=red]
            plot coordinates {
            (2,276)
            (3,132)
            (4,309)
            (5,274)
            (6,325)
            (7,416)
            (8,485)
            (9,418)
            (10,348)
            (11,271)
            (12,246)
            (13,154)
            (14,108)
            (15,76)
            (16,28)
            (17,10)
            (18,7)
            (19,3)
            (21,2)
            };
        \addlegendentry{\swint}
    
        \addplot[smooth,color=black]
            plot coordinates {
            (2,292)
            (3,141)
            (4,308)
            (5,257)
            (6,321)
            (7,421)
            (8,463)
            (9,445)
            (10,338)
            (11,256)
            (12,244)
            (13,144)
            (14,104)
            (15,58)
            (16,30)
            (17,15)
            (18,6)
            (19,2)
            (21,1)
            };
        \addlegendentry{\resnet}
        \end{axis}
        \end{tikzpicture}
    \end{subfigure}
    \hfill\hfill
    %% PascalVOC
    \begin{subfigure}[b]{0.3\textwidth}
        \centering
         \begin{tikzpicture}[scale=0.7]
            \begin{axis}[
            xlabel=Shortest Path (\voc),
            yticklabel=\empty,
            xmin=0, xmax=24,
            ymin=0, ymax=800,
            ]
        \addplot[smooth,blue] plot coordinates {
            (2,40)
            (3,23)
            (4,90)
            (5,94)
            (6,109)
            (7,116)
            (8,161)
            (9,165)
            (10,168)
            (11,131)
            (12,95)
            (13,58)
            (14,26)
            (15,32)
            (16,19)
            (17,13)
            (18,11)
            (19,6)
            (20,2)
            (21,1)  
        };
        \addlegendentry{\mob}
    
        \addplot[smooth,color=red]
            plot coordinates {
            (2,68)
            (3,40)
            (4,108)
            (5,121)
            (6,126)
            (7,90)
            (8,161)
            (9,175)
            (10,172)
            (11,109)
            (12,86)
            (13,46)
            (14,36)
            (15,43)
            (16,16)
            (17,18)
            (18,4)
            (19,6)
            };
        \addlegendentry{\swint}
    
        \addplot[smooth,color=black]
            plot coordinates {
            (2,75)
            (3,36)
            (4,106)
            (5,131)
            (6,128)
            (7,90)
            (8,170)
            (9,177)
            (10,155)
            (11,127)
            (12,71)
            (13,43)
            (14,22)
            (15,38)
            (16,14)
            (17,19)
            (18,8)
            (19,5)
            (21,1)
            };
        \addlegendentry{\resnet}
        \end{axis}
        \end{tikzpicture}
    \end{subfigure}
    \hfill
    \begin{subfigure}[b]{0.3\textwidth}
        \centering
         \begin{tikzpicture}[scale=0.7]
            \begin{axis}[
            xlabel=Shortest Path (\ecssd),
            yticklabel=\empty,
            xmin=0, xmax=24,
            ymin=0, ymax=800,
            ]
        \addplot[smooth,blue] plot coordinates {
            (2,37)
            (3,20)
            (4,59)
            (5,57)
            (6,67)
            (7,69)
            (8,141)
            (9,99)
            (10,105)
            (11,67)
            (12,71)
            (13,49)
            (14,56)
            (15,27)
            (16,19)
            (17,13)
            (18,1)
            (20,1)
        };
        \addlegendentry{\mob}

        \addplot[smooth,color=red]
            plot coordinates {
            (2,61)
            (3,28)
            (4,64)
            (5,62)
            (6,89)
            (7,77)
            (8,162)
            (9,111)
            (10,103)
            (11,58)
            (12,57)
            (13,40)
            (14,26)
            (15,21)
            (16,9)
            (17,8)
            (18,3)
            };
        \addlegendentry{\swint}
    
        \addplot[smooth,color=black]
            plot coordinates {
            (2,57)
            (3,28)
            (4,70)
            (5,77)
            (6,74)
            (7,95)
            (8,130)
            (9,103)
            (10,84)
            (11,84)
            (12,74)
            (13,32)
            (14,25)
            (15,15)
            (16,8)
            (17,7)
            (18,1)
            (19,1)
            (20,1)
            };
        \addlegendentry{\resnet}
        \end{axis}
        \end{tikzpicture}
    \end{subfigure}
    \caption{Shortest path between the original classification and its inverse in the \imagenet hierarchy over $3$ different datasets, with $\delta = 0.5$. There is remarkable similarity across all three models, a similarity which is consistent over the different datasets.}
    \label{fig:inverse_distance_05}
\end{figure*}

\begin{figure*}[t]
    \begin{subfigure}[b]{0.3\textwidth}
        \centering
         \begin{tikzpicture}[scale=0.7]
            \begin{axis}[
            xlabel=Shortest Path (\imagenet-1K),
            ylabel=No. Images,
            xmin=0, xmax=24,
            ymin=0, ymax=800,
            ]
        \addplot[smooth,blue] plot coordinates {
            (0,202)
            (2,122)
            (3,34)
            (4,144)
            (5,130)
            (6,246)
            (7,332)
            (8,425)
            (9,392)
            (10,413)
            (11,347)
            (12,334)
            (13,215)
            (14,154)
            (15,94)
            (16,50)
            (17,32)
            (18,9)
            (19,1)
            (20,1)
        };
        \addlegendentry{\mob}
    
        \addplot[smooth,color=red]
            plot coordinates {
            (0,145)
            (2,126)
            (3,43)
            (4,133)
            (5,115)
            (6,264)
            (7,376)
            (8,427)
            (9,441)
            (10,402)
            (11,353)
            (12,335)
            (13,282)
            (14,192)
            (15,124)
            (16,78)
            (17,30)
            (18,14)
            (19,4)
            (20,2)
            };
        \addlegendentry{\swint}
    
        \addplot[smooth,color=black]
            plot coordinates {
            (0,141)
            (2,130)
            (3,40)
            (4,158)
            (5,167)
            (6,288)
            (7,390)
            (8,425)
            (9,428)
            (10,412)
            (11,366)
            (12,309)
            (13,248)
            (14,155)
            (15,97)
            (16,49)
            (17,30)
            (18,9)
            (19,3)
            (22,1)
            };
        \addlegendentry{\resnet}
        \end{axis}
        \end{tikzpicture}
    \end{subfigure}
    \hfill\hfill
    \begin{subfigure}[b]{0.3\textwidth}
        \centering
         \begin{tikzpicture}[scale=0.7]
            \begin{axis}[
            xlabel=Shortest Path (\voc),
            yticklabel=\empty,
            xmin=0, xmax=24,
            ymin=0, ymax=800,
            ]
        \addplot[smooth,blue] plot coordinates {
            (0,41)
            (2,25)
            (3,8)
            (4,43)
            (5,54)
            (6,79)
            (7,109)
            (8,174)
            (9,143)
            (10,200)
            (11,146)
            (12,112)
            (13,84)
            (14,41)
            (15,40)
            (16,23)
            (17,15)
            (18,14)
            (19,5)
            (20,2)
            (21,1)
            (23,1)
        };
        \addlegendentry{\mob}
    
        \addplot[smooth,color=red]
            plot coordinates {
            (0,39)
            (2,30)
            (3,13)
            (4,56)
            (5,52)
            (6,79)
            (7,108)
            (8,158)
            (9,180)
            (10,179)
            (11,144)
            (12,110)
            (13,73)
            (14,59)
            (15,49)
            (16,46)
            (17,26)
            (18,12)
            (19,6)
            (20,4)
            (21,2)
            };
        \addlegendentry{\swint}
    
        \addplot[smooth,color=black]
            plot coordinates {
            (0,37)
            (2,27)
            (3,13)
            (4,61)
            (5,62)
            (6,74)
            (7,125)
            (8,172)
            (9,174)
            (10,198)
            (11,146)
            (12,102)
            (13,75)
            (14,44)
            (15,43)
            (16,29)
            (17,19)
            (18,11)
            (19,2)
            (21,2)
            };
        \addlegendentry{\resnet}
        \end{axis}
        \end{tikzpicture}
    \end{subfigure}
    \hfill
    \begin{subfigure}[b]{0.3\textwidth}
        \centering
         \begin{tikzpicture}[scale=0.7]
            \begin{axis}[
            xlabel=Shortest Path (\ecssd),
            yticklabel=\empty,
            xmin=0, xmax=24,
            ymin=0, ymax=800,
            ]
        \addplot[smooth,blue] plot coordinates {
            (0,46)
            (2,15)
            (3,3)
            (4,32)
            (5,28)
            (6,62)
            (7,66)
            (8,97)
            (9,97)
            (10,121)
            (11,83)
            (12,100)
            (13,67)
            (14,59)
            (15,34)
            (16,27)
            (17,12)
            (18,5)
            (19,3)
            (20,1)
        };
        \addlegendentry{\mob}

        \addplot[smooth,color=red]
            plot coordinates {
            (0,31)
            (2,23)
            (3,9)
            (4,27)
            (5,35)
            (6,64)
            (7,72)
            (8,111)
            (9,100)
            (10,99)
            (11,101)
            (12,92)
            (13,69)
            (14,55)
            (15,32)
            (16,28)
            (17,18)
            (18,10)
            (19,1)
            (20,1)
            (21,1)
            };
        \addlegendentry{\swint}
    
        \addplot[smooth,color=black]
            plot coordinates {
            (0,25)
            (2,35)
            (3,6)
            (4,34)
            (5,46)
            (6,76)
            (7,86)
            (8,115)
            (9,107)
            (10,115)
            (11,82)
            (12,84)
            (13,44)
            (14,41)
            (15,29)
            (16,17)
            (17,14)
            (18,6)
            (19,3)
            (21,1)
            };
        \addlegendentry{\resnet}
        \end{axis}
        \end{tikzpicture}
    \end{subfigure}
    \caption{Shortest path between the original classification and its adjustment pixels in the \imagenet hierarchy over $3$ different datasets, with $\delta = 0.5$. The distance between adjustment and target class on \imagenet-1K is obviously different from both \voc and \ecssd.}
    \label{fig:adjustment_distance_05}
\end{figure*}

\Cref{fig:adjustment_distance_05} shows the distance in the \imagenet hierarchy across $3$ datasets and $3$ models, given a $\delta = 0.5$ for \deltacomplete explanations, \ie the \deltacomplete explanation must contain at least $50\%$ of the original confidence.

\subsection{Caltech-256}

\begin{figure*}[t]
    \centering
    \scalebox{0.8}{\input{caltech_results.pgf}}
    \caption{Results on $3$ classes from Caltech-256. The general pattern here is the same as for the other datasets.}%
    \label{fig:caltech_results}
\end{figure*}

\begin{figure*}[t]
    \begin{subfigure}{0.4\textwidth}
        \centering
        \begin{tikzpicture}[scale=0.7]
            \begin{axis}[
            xlabel=Shortest Path,
            yticklabel=\empty,
            xmin=0, xmax=24,
            ymin=0, ymax=100,
            ]
        \addplot[smooth,blue] plot coordinates {
            (4,2)
            (5,7)
            (6,27)
            (7,21)
            (8,15)
            (9,64)
            (10,51)
            (11,37)
            (12,31)
            (13,19)
            (14,5)
            (15,7)
            (16,1)
            (17,1)
            (18,1)
            (19,1)
            (20,1)
            (21,1)
        };
        \addlegendentry{\mob}

        \addplot[smooth,color=red]
            plot coordinates {
            (2,1)
            (4,5)
            (5,13)
            (6,39)
            (7,10)
            (8,17)
            (9,58)
            (10,39)
            (11,36)
            (12,45)
            (13,17)
            (14,8)
            (15,7)
            (17,2)
            (18,2)
            };
        \addlegendentry{\swint}
    
        \addplot[smooth,color=black]
            plot coordinates {
            (4,6)
            (5,17)
            (6,50)
            (7,12)
            (8,19)
            (9,37)
            (10,45)
            (11,33)
            (12,40)
            (13,12)
            (14,9)
            (15,6)
            (16,1)
            (17,5)
            (18,4)
            (19,1)
            };
        \addlegendentry{\resnet}
        \end{axis}
        \end{tikzpicture}

    \caption{Shortest path between the original classification and its inverse in the \imagenet hierarchy over a subset of Caltech-256. There is remarkable similarity across all three models, a similarity which is consistent over the different datasets.}\label{fig:caltech_inverse_distance}
    \end{subfigure}
    \hfill
    \begin{subfigure}{0.4\textwidth}
        \centering
        \begin{tikzpicture}[scale=0.7]
            \begin{axis}[
            xlabel=Shortest Path,
            yticklabel=\empty,
            xmin=0, xmax=24,
            ymin=0, ymax=200,
            ]
        \addplot[smooth,blue] plot coordinates {
            (4,1)
            (5,2)
            (6,25)
            (7,71)
            (8,16)
            (9,58)
            (10,49)
            (11,13)
            (12,14)
            (13,9)
            (14,6)
            (15,5)
            (16,2)
        };
        \addlegendentry{\mob}

        \addplot[smooth,color=red]
            plot coordinates {
            (0,1)
            (3,1)
            (4,8)
            (5,1)
            (6,10)
            (7,8)
            (8,11)
            (9,123)
            (10,61)
            (11,19)
            (12,18)
            (13,14)
            (14,2)
            (15,4)
            (16,6)
            (17,3)
            (18,1)
            (19,1)
            };
        \addlegendentry{\swint}
    
        \addplot[smooth,color=black]
            plot coordinates {
            (4,9)
            (5,4)
            (6,20)
            (7,41)
            (8,22)
            (9,65)
            (10,60)
            (11,24)
            (12,14)
            (13,16)
            (14,10)
            (15,2)
            (16,1)
            (17,2)
            (18,6)
            };
        \addlegendentry{\resnet}
        \end{axis}
        \end{tikzpicture}

    \caption{Shortest path between the original classification and its adjustment pixels in the \imagenet hierarchy over a subset of caltech-256. There is a lot of similarity across all three models, a similarity which is consistent over the different datasets.}
    \label{fig:caltech_adjustment_distance}
    \end{subfigure}
    \caption{$\delta$-complete and adjustment distance on Caltech-256.}\label{fig:caltech}
\end{figure*}

\Cref{fig:caltech_results} shows a small study on $3$ different classes from Caltech-256~\citep{caltech256}. These are, in general, simple images compared to \imagenet. The general pattern seen in~\Cref{fig:results} does not change.

\Cref{fig:caltech} shows the complete and adjustment distance in the \imagenet hierarchy on a small study of $3$ different classes from Caltech-256. This follows a similar pattern to Figure 6 in the main paper.

\end{document}

%% file: preamble.tex
\usepackage[utf8]{inputenc}
\usepackage{pgfplots}
\DeclareUnicodeCharacter{2212}{−}
\usepgfplotslibrary{groupplots,dateplot}
\usetikzlibrary{patterns,shapes.arrows}
\pgfplotsset{compat=newest}

\usepackage{thmtools, thm-restate}

\usepackage{xspace}
\usepackage{amsmath}
\usepackage{mathtools}
\usepackage{amsthm}
\usepackage{nicefrac}
\usepackage{amsfonts}
\usepackage{pifont}
\usepackage{floatflt}
\usepackage{url}
\usepackage{dsfont}

\usepackage{tikz}
\usetikzlibrary{calc, positioning, shadows, backgrounds, arrows.meta, shapes.geometric}
\usepackage{multirow}
\usepackage{booktabs}
\usepackage{tabularx}
\newcolumntype{L}{>{\raggedright\arraybackslash}X}

\usepackage{algorithm}
\usepackage{algorithmic}

\theoremstyle{plain}
\newtheorem{theorem}{Theorem}[section]
\newtheorem{proposition}[theorem]{Proposition}
\newtheorem{lemma}[theorem]{Lemma}
\newtheorem{corollary}[theorem]{Corollary}
\theoremstyle{definition}
\newtheorem{definition}{Definition}

\theoremstyle{remark}

\theoremstyle{plain}

\makeatletter
\DeclareRobustCommand*\cal{\@fontswitch\relax\mathcal}
\makeatother

\newcommand{\xai}{\textsc{XAI}\xspace}
\newcommand{\gradcam}{\textsc{g}rad-\textsc{cam}\xspace}
\newcommand{\lime}{\textsc{lime}\xspace}

\newcommand{\rex}{\textsc{\sc r}e{\sc x}\xspace}

\newcommand{\imagenet}{ImageNet\xspace}

\newcommand{\ie}{\emph{i.e.}\xspace}
\newcommand{\eg}{\emph{e.g.}\xspace}

\newcommand{\commentout}[1]{}

\newcommand{\U}{{\cal U}}

\newcommand{\cF}{{\cal F}}
\newcommand{\V}{{\cal V}}

\newcommand{\K}{{\cal K}}

\newcommand{\sat}{\models}

\newcommand{\lem}{\begin{lemma}}
\newcommand{\elem}{\end{lemma}}
\newcommand{\pro}{\begin{proposition}}
\newcommand{\epro}{\end{proposition}}

\newcommand{\dfn}{\begin{definition}}
\newcommand{\edfn}{\end{definition}}

\newtheorem{example}{Example}
\newcommand{\xam}{\begin{example}}
\newcommand{\exam}{\end{example}}

\newcommand{\?}{\stackrel{?}{=}}

\usepackage{algorithm}
\usepackage{algorithmic}

\usepackage{longtable}
\usepackage{subcaption}
\usepackage{pgfplots}
\usepackage{pgfplotstable}
\usepackage{paralist}
\usepgfplotslibrary{fillbetween}
\pgfplotsset{compat=1.16}

\newcommand{\kplus}{$\mathcal{K}^+$\xspace}
\newcommand{\kminus}{$\mathcal{K}^{-}$\xspace}

%% file: inverse.tex
\begin{figure*}[t]
    \begin{subfigure}[b]{0.3\textwidth}
        \centering
         \begin{tikzpicture}[scale=0.7]
            \begin{axis}[
            xlabel=Shortest Path (\imagenet-1K),
            ylabel=No. Images,
            xmin=0, xmax=24,
            ymin=0, ymax=550,
            ]
        \addplot[smooth,blue] plot coordinates {
            (2, 164)
            (3, 76)
            (4, 206)
            (5, 221)
            (6, 270)
            (7, 396)
            (8, 496)
            (9, 449)
            (10, 380)
            (11, 342)
            (12, 293)
            (13, 220)
            (14, 150)
            (15, 107)
            (16, 48)
            (17, 24)
            (18, 10)
            (19, 1)
            (21, 1)
        };
        \addlegendentry{\mob}
    
        \addplot[smooth,color=red]
            plot coordinates {
            (2, 233)
            (3, 109)
            (4, 260)
            (5, 238)
            (6, 323)
            (7, 423)
            (8, 481)
            (9, 413)
            (10, 374)
            (11, 339)
            (12, 277)
            (13, 169)
            (14, 115)
            (15, 93)
            (16, 37)
            (17, 13)
            (18, 10)
            (19, 1)
            (20, 2)
            (21, 2)
            };
        \addlegendentry{\swint}
    
        \addplot[smooth,color=black]
            plot coordinates {
            (2, 262)
            (3, 140)
            (4, 284)
            (5, 244)
            (6, 320)
            (7, 436)
            (8, 456)
            (9, 443)
            (10, 350)
            (11, 271)
            (12, 277)
            (13, 170)
            (14, 115)
            (15, 66)
            (16, 30)
            (17, 16)
            (18, 9)
            (19, 3)
            (20, 1)
            (21, 1)
            };
        \addlegendentry{\resnet}
        \end{axis}
        \end{tikzpicture}
    \end{subfigure}
    \hfill\hfill
    %% PascalVOC
    \begin{subfigure}[b]{0.3\textwidth}
        \centering
         \begin{tikzpicture}[scale=0.7]
            \begin{axis}[
            xlabel=Shortest Path (\voc),
            yticklabel=\empty,
            xmin=0, xmax=24,
            ymin=0, ymax=550,
            ]
        \addplot[smooth,blue] plot coordinates {
            (2, 31)
            (3, 25)
            (4, 71)
            (5, 79)
            (6, 84)
            (7, 124)
            (8, 177)
            (9, 191)
            (10, 194)
            (11, 142)
            (12, 118)
            (13, 59)
            (14, 32)
            (15, 39)
            (16, 25)
            (17, 15)
            (18, 14)
            (19, 5)
            (20, 3)
            (21, 1)
        };
        \addlegendentry{\mob}
    
        \addplot[smooth,color=red]
            plot coordinates {
            (2, 57)
            (3, 40)
            (4, 106)
            (5, 113)
            (6, 117)
            (7, 98)
            (8, 162)
            (9, 190)
            (10, 155)
            (11, 130)
            (12, 88)
            (13, 44)
            (14, 41)
            (15, 38)
            (16, 32)
            (17, 22)
            (18, 5)
            (19, 3)
            };
        \addlegendentry{\swint}
    
        \addplot[smooth,color=black]
            plot coordinates {
            (2, 68)
            (3, 35)
            (4, 110)
            (5, 139)
            (6, 133)
            (7, 92)
            (8, 163)
            (9, 184)
            (10, 164)
            (11, 123)
            (12, 79)
            (13, 37)
            (14, 25)
            (15, 32)
            (16, 20)
            (17, 22)
            (18, 7)
            (19, 6)
            (21, 1)
            };
        \addlegendentry{\resnet}
        \end{axis}
        \end{tikzpicture}
    \end{subfigure}
    \hfill
    \begin{subfigure}[b]{0.3\textwidth}
        \centering
         \begin{tikzpicture}[scale=0.7]
            \begin{axis}[
            xlabel=Shortest Path (\ecssd),
            yticklabel=\empty,
            xmin=0, xmax=24,
            ymin=0, ymax=550,
            ]
        \addplot[smooth,blue] plot coordinates {
            (2, 27)
            (3, 12)
            (4, 56)
            (5, 54)
            (6, 65)
            (7, 76)
            (8, 127)
            (9, 111)
            (10, 112)
            (11, 76)
            (12, 83)
            (13, 61)
            (14, 60)
            (15, 24)
            (16, 23)
            (17, 16)
            (18, 3)
            (19, 2)
        };
        \addlegendentry{\mob}

        \addplot[smooth,color=red]
            plot coordinates {
            (2, 56)
            (3, 22)
            (4, 55)
            (5, 65)
            (6, 77)
            (7, 89)
            (8, 154)
            (9, 108)
            (10, 117)
            (11, 58)
            (12, 70)
            (13, 45)
            (14, 29)
            (15, 29)
            (16, 11)
            (17, 8)
            (18, 4)
            (19, 1)
            };
        \addlegendentry{\swint}
    
        \addplot[smooth,color=black]
            plot coordinates {
            (2, 53)
            (3, 25)
            (4, 65)
            (5, 72)
            (6, 83)
            (7, 101)
            (8, 132)
            (9, 106)
            (10, 93)
            (11, 89)
            (12, 77)
            (13, 37)
            (14, 27)
            (15, 16)
            (16, 10)
            (17, 7)
            (18, 1)
            (19, 1)
            (20, 1)
            };
        \addlegendentry{\resnet}
        \end{axis}
        \end{tikzpicture}
    \end{subfigure}
    \caption{Shortest path between the original classification and its inverse in the \imagenet hierarchy over $3$ different datasets. There is remarkable similarity across all three models, a similarity which is consistent over the different datasets.}
    \label{fig:inverse_distance}
\end{figure*}

%% file: completeness.tex
\begin{figure*}[t]
    \begin{subfigure}[b]{0.3\textwidth}
        \centering
         \begin{tikzpicture}[scale=0.7]
            \begin{axis}[
            xlabel=Shortest Path (\imagenet-1K),
            ylabel=No. Images,
            xmin=0, xmax=24,
            ymin=0, ymax=550,
            ]
        \addplot[smooth,blue] plot coordinates {
            (0,12)
            (2,17)
            (3,10)
            (4,38)
            (5,132)
            (6,382)
            (7,532)
            (8,418)
            (9,501)
            (10,461)
            (11,402)
            (12,384)
            (13,201)
            (14,136)
            (15,62)
            (16,29)
            (17,13)
            (18,4)
        };
        \addlegendentry{\mob}
    
        \addplot[smooth,color=red]
            plot coordinates {
            (0,50)
            (2,62)
            (3,22)
            (4,59)
            (5,82)
            (6,262)
            (7,458)
            (8,438)
            (9,466)
            (10,399)
            (11,408)
            (12,339)
            (13,306)
            (14,253)
            (15,114)
            (16,73)
            (17,23)
            (18,9)
            };
        \addlegendentry{\swint}
    
        \addplot[smooth,color=black]
            plot coordinates {
            (0,66)
            (2,59)
            (3,21)
            (4,82)
            (5,151)
            (6,378)
            (7,502)
            (8,444)
            (9,504)
            (10,449)
            (11,378)
            (12,324)
            (13,221)
            (14,140)
            (15,79)
            (16,47)
            (17,23)
            (18,14)
            (19,4)
            (22,1)
            };
        \addlegendentry{\resnet}
        \end{axis}
        \end{tikzpicture}
    \end{subfigure}
    \hfill\hfill
    \begin{subfigure}[b]{0.3\textwidth}
        \centering
         \begin{tikzpicture}[scale=0.7]
            \begin{axis}[
            xlabel=Shortest Path (\voc),
            yticklabel=\empty,
            xmin=0, xmax=24,
            ymin=0, ymax=550,
            ]
        \addplot[smooth,blue] plot coordinates {
            (0,2)
            (2,2)
            (3,2)
            (4,10)
            (5,44)
            (6,84)
            (7,230)
            (8,181)
            (9,189)
            (10,219)
            (11,130)
            (12,94)
            (13,71)
            (14,62)
            (15,47)
            (16,17)
            (17,12)
            (18,7)
            (19,2)
            (20,4)
            (21,1)
            (23,1)
        };
        \addlegendentry{\mob}
    
        \addplot[smooth,color=red]
            plot coordinates {
            (0,10)
            (2,9)
            (3,8)
            (4,20)
            (5,40)
            (6,80)
            (7,177)
            (8,129)
            (9,218)
            (10,200)
            (11,129)
            (12,89)
            (13,81)
            (14,58)
            (15,61)
            (16,61)
            (17,43)
            (18,7)
            (19,3)
            (20,3)
            (21,1)
            };
        \addlegendentry{\swint}
    
        \addplot[smooth,color=black]
            plot coordinates {
            (0,7)
            (2,12)
            (3,2)
            (4,28)
            (5,49)
            (6,99)
            (7,222)
            (8,179)
            (9,198)
            (10,198)
            (11,116)
            (12,78)
            (13,75)
            (14,54)
            (15,52)
            (16,33)
            (17,20)
            (18,11)
            (19,3)
            (21,3)
            };
        \addlegendentry{\resnet}
        \end{axis}
        \end{tikzpicture}
    \end{subfigure}
    \hfill
    \begin{subfigure}[b]{0.3\textwidth}
        \centering
         \begin{tikzpicture}[scale=0.7]
            \begin{axis}[
            xlabel=Shortest Path (\ecssd),
            yticklabel=\empty,
            xmin=0, xmax=24,
            ymin=0, ymax=550,
            ]
        \addplot[smooth,blue] plot coordinates {
            (0,2)
            (2,3)
            (3,4)
            (4,11)
            (5,20)
            (6,95)
            (7,116)
            (8,90)
            (9,117)
            (10,114)
            (11,93)
            (12,149)
            (13,66)
            (14,38)
            (15,32)
            (16,8)
            (17,12)
            (18,3)
            (19,3)
        };
        \addlegendentry{\mob}

        \addplot[smooth,color=red]
            plot coordinates {
            (0,10)
            (2,6)
            (3,12)
            (4,8)
            (5,38)
            (6,46)
            (7,92)
            (8,119)
            (9,111)
            (10,98)
            (11,97)
            (12,92)
            (13,71)
            (14,91)
            (15,37)
            (16,29)
            (17,21)
            (18,10)
            (19,3)
            (20,2)
            (21,1)
            };
        \addlegendentry{\swint}
    
        \addplot[smooth,color=black]
            plot coordinates {
            (0,7)
            (2,18)
            (3,3)
            (4,22)
            (5,46)
            (6,95)
            (7,130)
            (8,111)
            (9,119)
            (10,104)
            (11,94)
            (12,97)
            (13,43)
            (14,46)
            (15,25)
            (16,18)
            (17,7)
            (18,7)
            (19,2)
            };
        \addlegendentry{\resnet}
        \end{axis}
        \end{tikzpicture}
    \end{subfigure}
    \caption{Shortest path between the original classification and its adjustment pixels in the \imagenet hierarchy over $3$ different datasets. The distance between adjustment and target class on \imagenet-1K is obviously different from both \voc and \ecssd.}
    \label{fig:adjustment_distance}
\end{figure*}